%% file: main.tex
\documentclass[11pt]{article}
\usepackage[utf8]{inputenc}
\usepackage[T1]{fontenc}
\usepackage{times}
\usepackage{amsmath}
\usepackage{amssymb}
\usepackage[margin=1in]{geometry}
\usepackage{graphicx}
\usepackage[table]{xcolor}
\usepackage{booktabs}
\usepackage{enumitem}
\usepackage{multirow}
\usepackage{float}
\usepackage{array}
\usepackage{longtable}
\usepackage{caption}
\usepackage{subcaption}
\usepackage{natbib}
\setcitestyle{authoryear}
\usepackage{soul}
\usepackage{listings}
\usepackage{algorithm}
\usepackage{algpseudocode}
\usepackage{url}
\usepackage{hyperref}
\hypersetup{hidelinks,
  pdftitle={Engineering Simplicity: Simple Mechanism Interfaces Steer LLM Agents},
  pdfauthor={Kehang Zhu, Anand Shah, David C. Parkes}}
\usepackage{color-edits}
\addauthor{ma}{blue}

\providecommand{\Description}[2][]{}

\newcount\Comments  
\newcommand{\kibitz}[2]{\ifnum\Comments=1{\color{#1}{#2}}\fi}

\newif\ifshowrev \showrevfalse
\DeclareRobustCommand{\rev}[1]{\ifshowrev{\color{red}#1}\else#1\fi}   
\newenvironment{revblock}{\ifshowrev\color{red}\fi}{}                 
\newcommand{\revfloat}{\ifshowrev\color{red}\captionsetup{textfont={color=red},labelfont={color=red}}\fi} 
\makeatletter
\newcommand{\applabel}[2]{\phantomsection\edef\@currentlabel{#1}\label{#2}}  
\makeatother

\newcount\CommentsAdd  
\newcommand{\kibitzAdd}[2]{\ifnum\CommentsAdd=1{\color{#1}{#2}}\fi}
\definecolor{english}{rgb}{0.0, 0.5, 0.0}
\definecolor{tw}{rgb}{0.0, 0.0, 0.5}

\title{Engineering Simplicity:\\ \large \rev{Simple Mechanism Interfaces Steer LLM Agents}\thanks{%
  \footnotesize Thanks to Parker Whitfill and Kobi Gal for their helpful feedback. Authors' contact information, code, and data will be available at \url{https://github.com/KeHang-Zhu/
engineer_simplicity}. $^{*}$Kehang Zhu and Anand V.\ Shah contributed equally to this work.}}
\author{
  Kehang Zhu$^{*}$\\ Harvard University
  \and
  Anand Shah$^{*}$\\ MIT
  \and
  David C. Parkes\\ Harvard University
}
\date{\today}

\begin{document}
\maketitle

\begin{abstract}
\begin{revblock}
Can interaction formats and textual scaffolds help large language model (LLM) agents make better decisions, and do better decisions come with better explanations? We study these questions in auctions and matching, multi-agent environments with explicit rules and known optimal strategies. These settings let us vary how a decision problem is presented while retaining a benchmark for evaluating behavior. Drawing on human-motivated theories of simplicity, we compare interfaces that elicit a complete bid or ranking with sequential interfaces that make safe choices easier to identify. We then hold the interaction format fixed and vary reasoning scaffolds and rule descriptions. Across four model families, the ascending auction interface substantially reduces bid deviations. The matching comparison also shows why sequential responses require different error accounting from complete rankings. Laying out payoff contingencies and explaining why truth-telling is safe also improve choices, whereas prompts to plan through matching rounds or form beliefs about opponents worsen play overall. In auctions, these behavioral gains are not accompanied by corresponding improvements in measured verbal indicators of strategic understanding in the agents' short stated plans. Other prompts change those indicators without improving bids. Our findings suggest that human-motivated theories of simplicity can inform the design of decision environments for artificial agents. They also show why scaffolds should be evaluated through realized choices as well as explanations: improvements in one need not appear in the other.
\end{revblock}
\end{abstract}
\newpage
\setcounter{tocdepth}{2}

\section{Introduction} \label{sec:introduction}

\begin{revblock}
An LLM agent's decisions depend on the environment in which it is asked to act: which choices it sees, the order in which it makes them, and how the rules are explained. This creates a design question for systems that delegate decisions to LLMs. Can we improve their choices by changing the interaction format or providing a reasoning scaffold, and how should we tell whether that support worked? A model's explanation is one possible signal, but explanations need not faithfully report the computation behind an answer \citep{turpin2023language, lanham2023measuring}. We therefore evaluate both the decisions an interface elicits and the strategic understanding expressed in the agent's stated plan.

We use auctions and matching as a controlled multi-agent testbed. An auction allocates an item among competing bidders; a matching procedure assigns participants to opportunities, such as students to school seats. A \emph{mechanism} specifies how participants' actions determine allocations and, when relevant, payments. In the mechanisms studied here, an agent has a known optimal strategy regardless of how others act: bid its true value in the auction, or report its true preferences in matching. This property, called \emph{strategy-proofness}, makes truthful play a behavioral benchmark that does not require estimating opponents' policies. We can therefore ask whether a scaffold helps an agent follow a sound strategy, even when the scaffold encourages it to reason about a multi-agent problem.

Economic theories of simplicity offer a principled way to design this support. Human participants often deviate from truthful play even when a mechanism is strategy-proof \citep{kagel1993independent, kagel1995individual, rees2018suboptimal}. \emph{Obvious strategy-proofness} (OSP), introduced by \citet{li2017obviously}, strengthens the guarantee by taking the interaction itself into account: at any point where an agent could depart from the prescribed strategy, the worst outcome from following it is at least as good as the best outcome from departing. Unlike ordinary strategy-proofness, this comparison does not require the agent to match outcomes across separate hypothetical cases about others' actions. OSP is thus an influential bridge between an incentive guarantee and the reasoning needed to recognize it. Related theories examine demands on forward planning and on beliefs about others' beliefs \citep{borgers2019strategically, pycia2023theory, li2024designing}. We use these human-motivated ideas to choose interface changes and textual scaffolds for LLM agents.

We study two mechanisms across four model families: a second-price sealed-bid (SPSB) auction and student-proposing deferred acceptance (DA). Presentation varies along two dimensions. The first is the \emph{extensive form}: who acts when, what choices are available, and what information is visible at each choice. We compare a sealed bid with an ascending clock, and a submitted school ranking with a sequence of yes/no questions, holding the intended allocation rule fixed. The second dimension is the text accompanying a fixed interaction format: scaffolds for contingent reasoning, forward planning, and belief formation, and descriptions that state why truthful play is safe. For auctions we measure signed bid deviations from value and mean absolute deviations; for matching we measure departures from truthful preferences. We also score the auction plans for verbal indicators of strategic understanding. The matching explanations are logged but are not part of this text analysis.

The auction makes the distinction between rules and interfaces concrete. In an SPSB auction, the highest bidder wins and pays the second-highest bid. Bidding one's value is a \emph{dominant strategy}, meaning that no alternative bid gives a higher payoff for any fixed set of opponents' bids \citep{vickrey1961counterspeculation}. Yet recognizing this requires separating the chance of winning from the price paid. An ascending clock instead presents a rising price and asks only whether to stay or exit. Staying until one's value is reached makes the safety of the strategy visible at the point of choice. \citet{zhu2024evidence} previously studied LLM bidding under these auction formats. We test the comparison across GPT-4o, Claude, Gemini, and Gemma.

Matching provides a different decision problem. Under student-proposing DA, students rank schools; each school tentatively holds its highest-priority applicants, and rejected applicants are considered at their next choices until no further rejections occur \citep{gale1962college}. Truthful ranking is a dominant strategy for students \citep{roth1982economics}. An OSP implementation is not available for arbitrary school priorities \citep{ashlagi2018stable}; we study priority orders satisfying the acyclicity condition of \citet{ergin2002efficient}, for which it is available. We compare submitting a full ranking with an iterative implementation that asks whether a particular available school is the student's favorite.\footnote{We use \emph{standard DA} for the implementation that elicits complete rankings and \emph{iterative DA} for the sequential-query implementation.} All four model families misreport under standard DA. Iterative DA supplies a second interface comparison, but its partial reports require decision-level scoring: a zero distance between recorded rankings does not establish truthful responses (Section~\ref{sec:osp-da}).

Changing the interaction format may be impractical, so we also test support delivered entirely through text. Following \citet{li2024designing}, we distinguish reasoning about possible outcomes under others' actions, planning through later steps, and forming beliefs about other participants. The \emph{Payoff Tree} scaffold lays out the possible outcomes; planning prompts ask agents to trace DA's rounds; belief prompts ask them to consider opponents' actions or beliefs. Separately, work on descriptions that expose incentive properties \citep{gonczarowski2023strategyproofness, gonczarowski2024describing} motivates brief safety statements: an auction bid determines whether one wins, not the price paid; in DA, a rejection redirects an application to the next choice. These interventions let us compare explicit descriptions of a useful rule property with prompts that ask for additional reasoning.

Presentation changes behavior. The payoff-tree scaffold roughly halves deviations from truthful play in both domains. A safety description reduces mean error in complete matching rankings from $4.2\%$ to $0.2\%$. Planning and belief scaffolds instead worsen play overall. The auction plans reveal a further distinction: the successful payoff and safety interventions do not produce corresponding gains in our measured verbal indicators of strategic understanding. A worst-case scaffold changes what plans say without a detectable pooled bid improvement, while some harmful prompts change both language and bids. These results concern measured features of short explanations, not the models' unobserved reasoning processes.

The paper makes two contributions.
\begin{itemize}
    \item \textbf{A theory-guided approach to designing LLM decision environments.} Human-motivated simplicity principles identify changes to mechanism interfaces and descriptions that improve LLM choices across two economic settings and four model families. They also identify reasoning prompts that can be counterproductive.
    \item \textbf{An evaluation of behavioral and verbal responses to the same interventions.} In the auction testbed, improved choices need not be accompanied by improved verbal indicators of strategic understanding. Explanations can therefore complement, but cannot substitute for, direct evaluation of decisions.
\end{itemize}

Our focus is the design of environments for artificial agents. Work using LLMs as simulated human participants \citep{horton2023large, aher2023using, manning2024automated} and work studying their auction behavior \citep{chen2023put, fish2024algorithmic} provide related evidence, but our question is whether principles developed for human reasoners help us design effective support for LLMs. The emphasis on the difficulty of the decision problem also connects to \citet{oprea2024decisions}. The economic testbed gives this broader question a precise behavioral benchmark \citep{qian2025strategic}; it does not establish that the same interventions will improve unrelated tasks. Appendix~\ref{sec:appendix-related} discusses further related work.

Section~\ref{sec:methods} introduces the tasks and measurements. Section~\ref{sec:osp-results} tests interaction formats, and Section~\ref{sec:intervention-results} tests textual support within a fixed format. Section~\ref{sec:traces} compares changes in auction choices with changes in stated plans. Section~\ref{sec:conclusion} discusses implications for the design and evaluation of LLM decision environments.
\end{revblock}

\section{Experimental Design} \label{sec:methods}

\rev{Auctions and matching are canonical multi-agent allocation problems: auctions assign goods using bids and payments, whereas matching assigns participants to opportunities using preferences and priorities. In both settings studied here, truthful reporting is a dominant strategy, giving us a common benchmark for evaluating whether an interface helps an agent choose well. We first define the tasks, then describe how we elicit decisions and measure departures from this benchmark.}

\subsection{Mechanisms}

We study two strategy-proof mechanisms, each paired with an OSP implementation for the environments below. \rev{The mechanism specifies how actions translate into outcomes; the interface specifies how those actions are elicited. The comparison changes the interface while preserving the intended allocation rule.}

\paragraph{Auctions.} We implement a second-price sealed-bid (SPSB) auction with $N=3$ bidders and independent private values, alongside an ascending clock auction that implements the same allocation and payment rule. Human subjects have historically deviated from truthful bidding in SPSB auctions \citep{kagel1993independent, kagel1995individual}, while ascending clock formats improve play \citep{li2017obviously, breitmoser2022obviousness}. Details of the auction environment appear in Section~\ref{sec:auction-data}.

\paragraph{Matching.} We implement student-proposing deferred acceptance (DA) in a school choice setting with four students and four schools, alongside an iterative DA protocol. We construct priority structures that are Ergin-acyclic---ensuring the iterative protocol is OSP---but with non-aligned preferences, so the matching problem is non-trivial. Details of the matching environment appear in Section~\ref{sec:da-data}.

\subsection{Data Generation for Deferred Acceptance} \label{sec:da-data}
\paragraph{Overview.}
We generate data from a school choice environment in which LLMs play the role of students. Each experimental repetition instantiates a complete matching market (student preference rankings, and school priorities), elicits student reports under a specified mechanism, and  executes student-proposing Deferred Acceptance (DA) to produce a
matching outcome.

\paragraph{Market instances.}
Each instance contains students $i\in\{A,B,C,D\}$ and schools $s\in\{w,x,y,z\}$, with unit capacity (one seat per school). \rev{Student values follow an affiliated model, $v_{i,s} = c_s + \varepsilon_{i,s}$, with a school-level common component $c_s \sim \mathrm{Unif}[40,70]$ and a student--school shock $\varepsilon_{i,s}\sim \mathrm{Unif}[0,20]$, so that demand is correlated across students while tastes remain idiosyncratic; each student's true preference ranking is the descending sort of her values \citep{klijn2019static}. Schools rank students by fixed, Ergin-acyclic priority orders that create a top tier $\{A,B\}$ and a bottom tier $\{C,D\}$---the condition under which a sequential-query implementation of DA is OSP---and all students observe a fixed ``global popularity'' ranking ($y \succ x \succ w \succ z$) that proxies common beliefs about demand and may or may not align with their own preferences. The exact priority orders and protocol details are in Appendix~\ref{sec:appendix-details}.}

\paragraph{Mechanisms.}
For each market instance we run two elicitation protocols. \emph{(i) Direct revelation (static DA).} Each student submits a complete rank-order list over $\{w,x,y,z\}$ in a single shot. We then run student-proposing DA on the submitted rankings and with the true school priorities. \emph{(ii) Iterative DA (sequential queries).} Students do not submit a full ranking upfront. Instead, the mechanism queries students sequentially, following the tree construction of \citet{ashlagi2018stable} for OSP implementation of DA under acyclic priorities. \rev{At each node a student holding top priority at some remaining school is asked a yes/no question---``Among the remaining schools, is school $s$ your most preferred?''---and is matched immediately on YES; the protocol lets the student choose her favorite remaining school when a single top-priority student remains, and terminates when all students are matched.}

\paragraph{LLM elicitation as scenario prompts.}
Within each repetition, we query one LLM instance per student. Each query is a self-contained prompt that bundles: (a) the student's identity,  (b) their priority  at each school, (c) the global ranking signal, (d) the mechanism rules, and (e) a treatment-specific reasoning scaffold (if applicable). We elicit a written explanation before each decision, in the spirit of reasoning prompts studied by \citet{wei2022chain}, using separate tags for the explanation and action. \rev{Throughout, we refer to the text inside the reasoning tag as the agent's \emph{stated plan}: a short explanation produced before acting, not a complete record of the model's computation (Section~\ref{sec:traces}).} \rev{Prompt skeletons for both protocols are in Appendix~\ref{sec:appendix-prompts}.}

\paragraph{Repetitions and logging.}
For each treatment condition and each mechanism, we target $N=50$ independent repetitions (distinct random seeds for value draws; priorities and global ranking fixed)\rev{, logging true rankings, raw LLM responses, parsed decisions, DA traces, and final matches so that outcomes can be replayed and audited; retained counts can be lower after response or parsing failures (Appendix~\ref{sec:appendix-details})}.

\subsection{Data Generation for Auctions} \label{sec:auction-data}

\paragraph{Overview.}
We generate auction data from an in-silico laboratory in which large language models (LLMs) act as bidders. 
Each experimental repetition instantiates a complete one-shot  auction environment (values, mechanism, and intervention), elicits a sequence of actions from each bidder (a sealed bid or a sequence of stay/exit decisions), and then computes allocations and payments mechanically from the submitted actions. 
The unit of observation is an \emph{auction instance}: a fully specified scenario that is reproducible given a random seed and a prompt template.

\paragraph{Environment and mechanisms.}
We study symmetric independent-private-value (IPV) auctions with $N=3$ bidders\rev{; each bidder's value is drawn independently and uniformly from $\{0,1,\dots,49\}$, and bids lie on a \$0.01 grid (\$0.50 for clock prices). In the \emph{second-price sealed-bid} (SPSB) auction, bidders submit bids simultaneously; the highest bidder wins and pays the second-highest bid. In the \emph{ascending clock} auction, the price rises from zero in fixed increments and each active bidder chooses at every posted price whether to stay or exit, without being told when others leave; the last bidder standing wins at the price at which the second-to-last bidder exited. Both formats implement the same allocation and payment rule; formal definitions are in Appendix~\ref{sec:appendix-details}.}

\paragraph{LLM elicitation as scenario prompts.}
In both mechanisms, each bidder’s decision is elicited via a self-contained prompt that includes: (a) bidder identity and opponent list, (b) a mechanism description (which varies by intervention), (c) the bidder’s private value $v_i$, and (d) a structured response format. \rev{The \texttt{<PLAN>} block (requested maximum of 50 words; realized median 68 words) is retained verbatim for every bid and forms the corpus analyzed in Section~\ref{sec:traces}.} \rev{Prompt skeletons for both formats are in Appendix~\ref{sec:appendix-prompts}.}

\paragraph{Treatments (prompt interventions).}
Treatments modify only the rule-explanation block of the prompt while holding the value distribution, bidder identities, and payment rules fixed. We use this to test how different reasoning scaffolds change bidding behavior under the same underlying \rev{game} (e.g., contingent-reasoning prompts, decision-tree framing, backward-induction analogies, direct strategy revelation, and risk-preference framings). \rev{Beyond the interventions reported in Section~\ref{sec:intervention-results}, two further prompts enter the trace analysis of Section~\ref{sec:traces}: a \emph{worst-case} contingent-reasoning scaffold, which asks the bidder to reason about the worst case for each bid rather than laying the cases out, and a \emph{menu restatement} of the second-price rule in the spirit of \citet{gonczarowski2023strategyproofness}, which describes the winner's payment as a ``price to win'' set by the other bids. Their texts appear in Appendix~\ref{sec:appendix-prompts-auction}.}

\subsection{Models}

We test four models spanning three commercial API providers and one open-weight family: GPT-4o \citep{openai2024gpt4o}, Claude 3.5 Haiku \citep{anthropic2024claude}, Gemini 2.0 Flash \citep{google2024gemini}, and Gemma 3 27B \citep{google2025gemma3}. All experiments use temperature $T=.5$.

\rev{We select models that can parse the tasks but still make enough errors to study how interface design affects their choices. This is a study of the named models, not a ranking of current model capabilities. Appendix~\ref{sec:traces-implications} reports an additional panel of more capable models as a boundary condition: models already near truthful play leave little room for these interventions to help.}

\section{OSP Results} \label{sec:osp-results}

\begin{figure}[htbp]
    \centering
    \includegraphics[width= \textwidth]{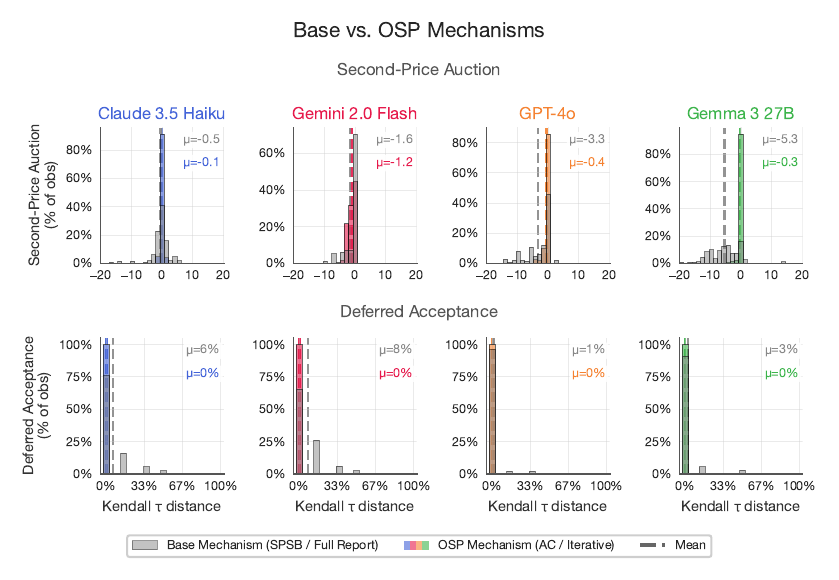}
    \caption{Auction and matching interface comparisons across four model families. \emph{Top:} signed bid deviations ($b-v$) in sealed-bid (gray) and ascending-clock (colored) auctions; dashed lines indicate means. \emph{Bottom:} recorded normalized Kendall $\tau$ distances in direct DA and iterative DA. Direct DA elicits full rankings; iterative DA reveals only partial preferences. Its zero recorded distance does not establish error-free responses: observations with fewer than two ranked schools contribute zero, and the statistic does not score all yes/no answers. Separate decision-level errors for the sequential-query protocol are reported in the text and Appendix~\ref{sec:appendix-osp-more}.}
    \label{fig:osp-comparison}
\end{figure}

\rev{We first ask whether breaking a decision into simpler sequential choices improves behavior. The comparisons preserve the intended allocation rule while changing the extensive form, or sequence of choices and information available to the agent. Ordinary strategy-proofness guarantees that truthful play is optimal; OSP additionally requires that, at each point where the prescribed strategy and a deviation first diverge, the \emph{worst} outcome from following the strategy is at least as good as the \emph{best} outcome from deviating \citep{li2017obviously}. This makes the guarantee recognizable without comparing each truthful outcome with a separate counterfactual under the same opponents' actions. Figure~\ref{fig:osp-comparison} tests whether that difference helps LLMs in auctions and matching.}

\subsection{Auctions: SPSB vs.\ Ascending Clock} \label{sec:osp-auctions}

The top row of Figure~\ref{fig:osp-comparison} shows the distribution of bid deviations from value in SPSB auctions. All four models deviate from truthful bidding, with mean deviations of $\mu = -0.5$ (Claude), $-1.6$ (Gemini), $-3.3$ (GPT-4o), and $-5.3$ (Gemma). The negative sign indicates systematic {\em underbidding}---models bid below their values. This contrasts with the primary finding in human experiments, where subjects tend to {\em overbid} in second-price auctions \citep{kagel1993independent}. The distributions also exhibit long tails, which are evidence of very large errors: some bids deviate by \$10--20 from value, representing cases where the model substantially misplays the auction.

Yet, our results also show the ascending clock format corrects deviations toward truthful play across all models.
Mean deviations compress to $\mu = -0.1$ (Claude), $-1.2$ (Gemini), $-0.5$ (GPT-4o), and $-0.3$ (Gemma). The improvement is most dramatic for the models with the worst SPSB performance: Gemma's mean deviation falls from $-5.3$ to $-0.3$, a near-complete correction. \rev{The large deviations visible in the sealed-bid distributions largely disappear in these samples. This reduction in dispersion matters alongside the change in average bidding.}

\rev{Values are private and independently drawn, and our clock does not reveal opponents' dropout decisions. Its gains therefore arise without supplying a new signal about the bidder's own value.} This raises a natural question: OSP simultaneously simplifies play along multiple cognitive dimensions---it eliminates the need for contingent reasoning, reduces each decision to a simple binary choice (stay or exit at each price), and makes the connection between action and outcome immediate. Which of these simplifications actually drives the observed improvement in play? We return to this question in Section~\ref{sec:intervention-results}. \rev{The agents' own clock rationales also bear on it: Appendix~\ref{sec:traces-clock} shows that exit decisions track value equally well whether or not the agent states the ``stay while the price is below value'' rule.}

\subsection{Matching: Direct DA vs.\ Iterative DA} \label{sec:osp-da}

Student-proposing deferred acceptance is strategy-proof for the proposing side: truthful reporting of preferences is a dominant strategy \citep{roth1982economics}. Yet human subjects routinely misreport in practice. For example, \citet{rees2018suboptimal} documents suboptimal play in the National Resident Matching Program, finding that a nontrivial fraction of applicants submit rankings that are inconsistent with revealed preferences---despite the mechanism being well-known and the stakes substantial. DA is generally not OSP-implementable \citep{ashlagi2018stable}, but under Ergin-acyclic priorities \citep{ergin2002efficient}, an iterative query protocol can implement the proposer-optimal stable matching in an OSP manner. To our knowledge, no prior work has experimentally tested an OSP implementation of DA with either human or LLM subjects.

The bottom row of Figure~\ref{fig:osp-comparison} shows the distribution of normalized Kendall $\tau$ distance---the fraction of pairwise ranking swaps relative to the true preference order---in direct-revelation DA. All models exhibit nontrivial error rates: $\mu = 6\%$ (Claude), $8\%$ (Gemini), $1\%$ (GPT-4o), and $3\%$ (Gemma). \rev{These ranking errors are measured on a different scale from auction deviations, which are in dollars. A misreport can nevertheless affect other students: when one assignment changes, a rejection may send an applicant to another school and displace someone there.}

\rev{Iterative DA has zero recorded Kendall distance, but this statistic is not a complete measure of response accuracy. A student matched after one choice provides no pair of schools to rank, and yes/no errors need not appear in a reconstructed ranking. A separate analysis of the sequential-query logs counts errors in both binary responses and final choices, excluding forced final choices and tolerating equal-valued options. It finds $16/194$ errors for Claude ($8.2\%$), $0/179$ for Gemini, $0/188$ for GPT-4o, and $3/210$ for Gemma ($1.4\%$). Appendix~\ref{sec:appendix-osp-more} gives the denominators and error types. These decision-level rates and complete-ranking distances use different units, so the comparison does not establish a common-scale reduction in matching errors.}

\rev{The interface comparison thus gives clear evidence of improved auction bidding and a complementary matching test whose measurement depends on what the interface elicits. We next evaluate textual support while holding the interaction format, and hence the behavioral measure, fixed.}

\section{Intervention Results} \label{sec:intervention-results}

\rev{The auction interface improves behavior, but changing an interaction protocol may be costly or infeasible. We next ask whether textual support can help while agents still submit a single bid or ranking. Each treatment changes only the explanation accompanying that decision; the rules, value distributions, and response format are held fixed.}

\rev{Following \citet{li2024designing}, we organize reasoning scaffolds around three demands: considering possible outcomes under others' actions, planning through later steps, and forming beliefs about other participants. We also test descriptions that directly explain a property making truthful play safe \citep{gonczarowski2023strategyproofness}. The main text reports the payoff-tree scaffold and these safety descriptions. Forward planning is tested in matching, where a ranking is processed through rounds of applications and rejections; these results and the belief prompts are in Appendix~\ref{sec:appendix-results}. Additional prompts about risk and payoff framing appear in Appendix~\ref{sec:appendix-prospect}.}

\rev{For auctions, the figures report signed bid deviations: negative values indicate underbidding, and values near zero indicate less systematic bias. Section~\ref{sec:traces} also reports absolute deviations, which avoid cancellation between overbids and underbids. For standard DA, error is the fraction of school pairs ordered incorrectly; smaller values are better. Results are reported in the fixed model order Claude, Gemini, GPT-4o, Gemma.}

\subsection{Contingent Reasoning} \label{sec:contingent}

\citet{li2024designing} identifies contingent reasoning as the core cognitive demand of non-OSP mechanisms. To see that truth-telling is dominant in SPSB, a bidder must consider each possible profile of opponent bids and check that no deviation improves payoffs in any case---a ``case-by-case comparison, calculating payoffs for each profile of opponent bids'' \citep{li2024designing}. The ascending clock eliminates it: at each price, the worst case from following the dominant strategy weakly dominates the best case from any deviation, without reference to opponents. The question, then, is whether scaffolding for contingent reasoning---without changing the extensive form---can partially recover these gains.

\rev{Our \emph{Payoff Tree} intervention lays out the possible outcomes of an action in words. The auction version also states the distinction between winning and payment:} 
\begin{itemize}
    \item In auctions: ``PATH A: your bid exceeds others $\to$ you WIN, and pay the second-highest bid. PATH B: your bid falls below $\to$ you LOSE, and pay nothing. Key insight: your bid determines which path you are on, not the payment in Path A.''
    \item In DA: ``Think of DA as a tree of possibilities. ROOT: You propose to your 1st
      choice. If ACCEPTED: you are matched. If REJECTED: you propose to your 2nd choice.
      If ACCEPTED: you are matched. If REJECTED: you propose to your 3rd choice. And so on. At each branch, acceptance or rejection depends on the school's priorities and who else proposed there.''
\end{itemize}

\begin{figure}[htbp]
    \centering
    \includegraphics[width=\textwidth]{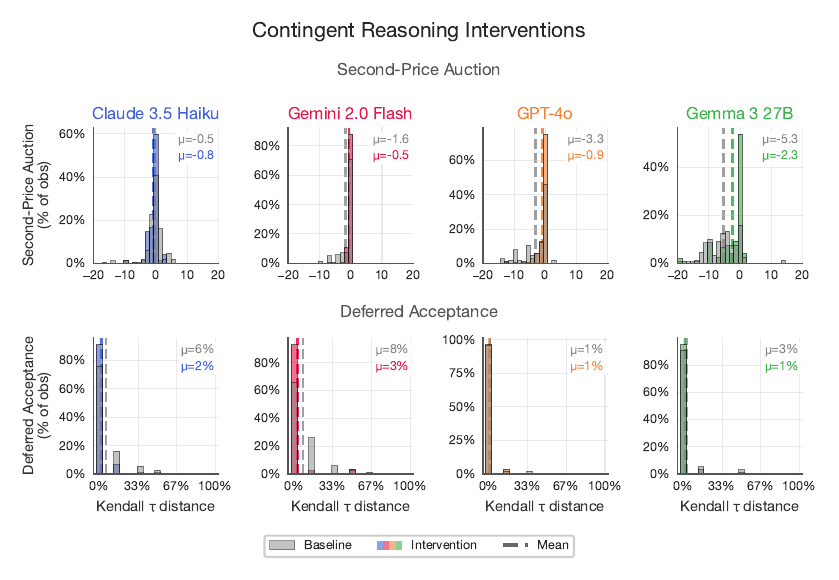}
    \caption{Contingent reasoning: \textit{Payoff Tree} intervention. {\em Top row:} Second-price auction bid deviations ($\text{bid} - \text{value}$); baseline SPSB in gray, intervention in color. {\em Bottom row:} Deferred acceptance Kendall $\tau$ errors; baseline direct DA in gray, intervention in color. The payoff tree scaffold consistently improves play across all models in both domains, roughly halving the distance from truthful.}
    \label{fig:contingent}
\end{figure}

In auctions, \textit{Payoff Tree} reduces the mean deviation from $\mu = -2.67$ to $\mu = -1.13$ across models, roughly halving the distance from truthful. The improvement is consistent across models: GPT-4o moves from $-3.28$ to $-0.86$, Gemma from $-5.27$ to $-2.34$, Gemini from $-1.63$ to $-0.53$. Claude, already close to truthful ($-0.49$), is the one model that shows null improvement. \rev{(In absolute deviation Claude improves as well, from $2.02$ to $1.11$ against the pooled baseline of Section~\ref{sec:traces}; the signed null masks a reduction in dispersion around value.)} For comparison, the ascending clock achieves $\mu = -0.5$ across models (Section~\ref{sec:osp-auctions}); the \textit{Payoff Tree} closes roughly half the gap between SPSB and OSP without changing the extensive form.

In DA, \textit{Payoff Tree} reduces mean error from $\mu = 4.2\%$ to $\mu = 1.7\%$ across models, again less than half the baseline error. Here the improvement is unanimous across all four models: Claude from $5.8\%$ to $2.0\%$, Gemini from $7.6\%$ to $2.8\%$, GPT-4o from $1.0\%$ to $0.6\%$, Gemma from $2.6\%$ to $1.4\%$. This comparison holds the complete-ranking interface fixed, unlike the sequential comparison in Section~\ref{sec:osp-da}.

\rev{These results show that explicitly describing payoff contingencies can improve choices without changing the interaction format. They do not isolate a single reasoning process: the auction scaffold also states the useful fact that the bid changes whether one wins rather than the payment. A related scaffold that merely asks the bidder to consider the \emph{worst case} for each bid changes what agents write but produces no detectable pooled improvement in bids (Section~\ref{sec:traces-probe}). The content of the support matters, beyond asking the model to reason more.}

\subsection{Mechanism Description} \label{sec:mech-description}

\rev{A different form of support explains a useful property of the rules directly. An incentive-compatible mechanism makes truthful reporting optimal, but that guarantee may be hard to infer from an algorithmic description. Work on ``strategyproofness-exposing'' descriptions asks how to make this guarantee easier to see \citep{gonczarowski2023strategyproofness}. Inspired by that approach, we test an \emph{outcome safety} description in each setting.}

The key step in \citet{vickrey1961counterspeculation}'s original argument for the strategy-proofness of the second-price auction comes from an argument about being `pivotal'. Namely, that your bid only determines \textit{that }you win, not \textit{what} you pay. This inspires our \emph{Payoff Safety} intervention for the auction setting: ``Think of your bid as setting your maximum price. Your bid determines IF you win, not WHAT you pay. If you win, you pay what others bid, not what you bid.''

Similarly, a key difference between two famous school choice mechanisms---the Boston mechanism \citep{abdulkadirouglu2003school} and the DA mechanism---is that under DA, rejections are safe. A rejection from a school does not hurt a student's chances at their next-ranked school under DA, whereas under the Boston mechanism, seats at other schools may fill while a student's application is being considered, making rejection costly. In practice, this difference was a key argument for replacing the Boston mechanism with DA in Boston Public Schools. Similarly, our \emph{Rejection Safety} intervention states: ``Rejections only redirect, never eliminate. Your ranking determines the \emph{order} in which you are considered, not \emph{whether} you are considered.''

Neither intervention scaffolds a particular mode of reasoning. They do not ask the agent to trace through contingencies, simulate the algorithm, or reason about opponents. They simply state the key property of the mechanism that makes it strategy-proof.

\begin{figure}[htbp]
    \centering
    \includegraphics[width=\textwidth]{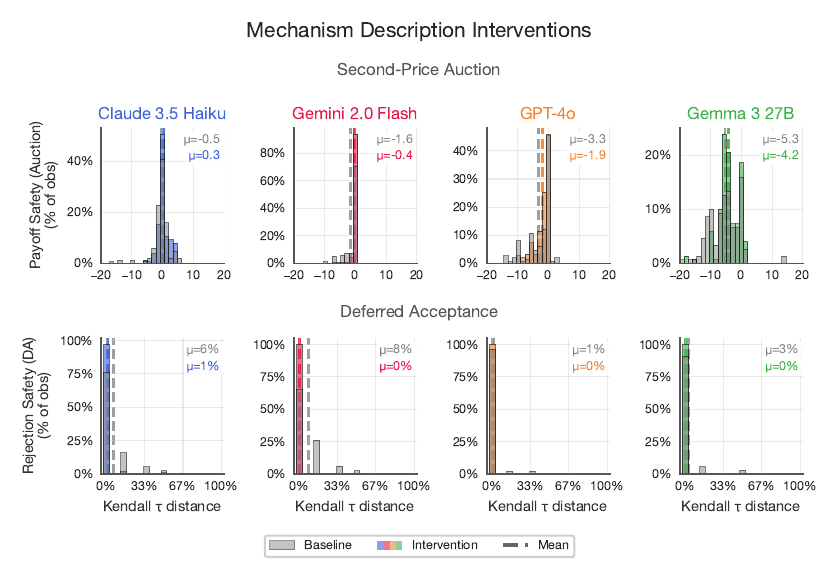}
    \caption{Mechanism description: outcome safety interventions. {\em Top row:} \textit{Payoff Safety} in second-price auctions (``your bid determines IF you win, not WHAT you pay''). {\em Bottom row:} \textit{Rejection Safety} in deferred acceptance (``rejections only redirect, never eliminate''). Baseline in gray, intervention in color. Both interventions improve play dramatically. Rejection safety reduces mean complete-ranking error in DA from $4.2\%$ to $0.2\%$.}
    \label{fig:mechanism-description}
\end{figure}

In DA, the results are even more striking. \textit{Rejection Safety} nearly eliminates error, reducing it from $\mu = 4.2\%$ to $0.2\%$ across all models. This is measured within the complete-ranking interface and should not be equated with the partial-ranking statistic of iterative DA.

\rev{A short description can therefore improve behavior substantially without changing the interaction format. This supports a practical role for explanations of incentive properties alongside changes to the interaction format. It does not establish which internal computation changed, or which component of OSP caused its benefit. Section~\ref{sec:traces} tests a narrower, observable question: whether better auction bids are accompanied by more frequent statements of the relevant strategic principles.}

\begin{revblock}
\section{What the Traces Say: Stated Reasoning and Realized Bids} \label{sec:traces}

The preceding sections measure whether agents make better choices. We now ask whether the same interventions improve what agents say about those choices. Every auction bid arrives with a \emph{stated plan}: a short explanation in the \texttt{<PLAN>} tag, written before acting (median 68 words). We compare treatment effects on bids with effects on observable features of these plans, such as stating the correct payment rule or explaining why truthful bidding is safe. The plans need not reveal the model's underlying computation \citep{turpin2023language, lanham2023measuring}; our question concerns the relationship between measured verbal indicators and behavior. DA also logs explanations, but they have not been scored in this analysis, so all results in this section concern auctions.

\subsection{Stated Plans as Data} \label{sec:traces-data}

\paragraph{Corpus and scoring.}
The corpus contains $21{,}990$ plans from the four model families of Section~\ref{sec:methods}: every sealed-bid cell (baseline and intervention), the ascending clock, and a set of first- and third-price cells recovered for GPT-4o that serve only as a cross-mechanism control. The analysis pool for all share and regression statistics is the $14{,}073$ sealed second-price plans; at temperature $0.5$ roughly $40\%$ of these are byte-identical duplicates within model, cell, and value draw, so we report every headline on the deduplicated pool ($n = 8{,}839$) as well.\footnote{Duplicates share a single value and a single bid, so full-sample standard errors are optimistic; every quantity below is stable under deduplication (Appendix~\ref{sec:appendix-traces}).} Each plan is scored with a frozen dictionary of eighteen binary keyword features---transparent regular expressions grouped into normative recognition (dominance language, correct payment-rule rehearsal, an echo of the payoff-safety invariant), opponent and belief reasoning, stated intent to bid below or above value, safety and risk rhetoric, and style---and is then assigned a primary \emph{decision script} by a rule-based decision list over those features (Appendix~\ref{sec:appendix-traces}). The dictionary was re-anchored on the vocabulary models actually produce (``overpay,'' ``profit margin,'' ``buffer'') after theory-native phrasings (``no risk,'' ``can't lose'') proved essentially absent: the feature that detects an echo of the \textit{Payoff Safety} invariant fires on $3$ of $21{,}990$ plans corpus-wide.

\paragraph{Validation.}
Because keyword rules can miss relevant meanings, an independent LLM judge re-labeled a stratified sample of $2{,}573$ plans, seeing only the plan text and the mechanism family (no bid, value, model, or rule label). Agreement with the rule-based label is $\kappa = 0.55$ at the coarse grain we use in the text---below-value script, opponent script, anchoring, aggressive, clock exit, or normative---and $\kappa = 0.36$ at the full ten-label grain, so script shares are reported only at the coarse grain and the confusion matrix goes to the appendix. The judge also finds little explicit safety reasoning under a broader definition: its broader ``states why truthful bidding is safe'' flag fires on $2.7\%$ of our panel's plans (and, as Appendix~\ref{sec:traces-implications} notes, on $45\%$ of frontier-model plans).

\paragraph{Plans are not noise.}
Where a plan names a dollar amount, the realized bid matches it almost exactly (correlation $0.97$--$0.99$ across the four families), and the stated direction---below, at, or above value---agrees with the realized direction for $83$--$98\%$ of plans. These associations suggest that many deviations are already present in the stated plan rather than introduced only when the numerical bid is produced. The modal plan in the sealed second-price auction is \emph{first-price} logic---``bid just below my value to keep a profit margin''---a script that would serve a bidder well where the winner pays her own bid, carried unchanged into a rule under which she does not; each family's plans show a characteristic mix of such scripts, and the plans that anchor on no value-based rule at all carry the largest errors. Appendix~\ref{sec:appendix-traces} documents the scripts with verbatim exemplars, the model fingerprints, and the fidelity analysis (Figures~\ref{fig:heuristic-prevalence} and~\ref{fig:fidelity}).

\paragraph{Measuring changes in language and bids.}
Each intervention now has two outcomes---the change in the prevalence of the language it targets, and the change in mean absolute deviation $|b-v|$---and we read the \emph{joint} response. Four contrasts are designated as primary (\textit{Payoff Safety}, \textit{Payoff Tree}, the worst-case scaffold, and the menu restatement of Section~\ref{sec:auction-data}) and are classified at their raw wild-cluster-bootstrap $p$-values, clustered at the run level. The remaining sixteen levers ---the belief scaffolds of Appendix~\ref{sec:beliefs}, the two-stage and clock-framed descriptions, and the prospect-theoretic personas and frames of Appendix~\ref{sec:appendix-prospect}---are exploratory and are classified at Benjamini--Hochberg $q$-values, with language tests and bid tests corrected as two separate families. We use two one-sided tests (TOST) to assess equivalence against pre-committed margins ($\pm\$0.50$ and $\pm\$1.00$ on $|b-v|$, $\pm 5$ percentage points on prevalence) rather than interpreting a nonsignificant difference as proof of no effect, and every bid effect is cross-checked with a four-model-cluster permutation test, which we flag as ``soft when it disagrees. The bid effects for 	extit{Payoff Safety}, 	extit{Payoff Tree}, first-order beliefs, and the risk-averse persona do not reach significance in that four-model check.\footnote{Two conventions coexist in the paper. Section~\ref{sec:intervention-results} reports signed mean deviation against the dedicated SPSB baseline cell (e.g., $-2.67 \to -1.53$ for \textit{Payoff Safety}). This section reports mean $|b-v|$ against the pooled baseline cells of the contingent-reasoning and belief grids ($n = 1{,}188$ plans), the cells that share the treated prompts' rule text; the corresponding \textit{Payoff Safety} contrast is $3.20 \to 1.95$. The two conventions agree in sign for every lever discussed; where they differ in emphasis (Claude under the \textit{Payoff Tree}; Gemini under the belief prompts) the text says so.}

\subsection{Intervention Effects on Language and Bids} \label{sec:traces-probe}

For each intervention, we compare changes in the prevalence of its targeted language with changes in mean absolute bid error. Figure~\ref{fig:language-bids} summarizes selected contrasts; Table~\ref{tab:dissociation} reports the main comparisons, and Appendix~\ref{sec:appendix-traces} reports all twenty interventions.

\paragraph{Primary contrasts.}
\textit{Payoff Safety} states the property that makes truthful bidding safe. Bids improve substantially---mean $|b-v|$ falls from $3.20$ to $1.95$ (wild-cluster $p = 0.015$), in all four families---yet the measured indicators of stated incentive understanding do not improve: the invariant is echoed in \emph{zero} of its $600$ treated plans (a rule-of-three upper bound of $0.5\%$), and correct payment-rule rehearsal is equivalent to no change within $\pm 5$ points. The measured verbal indicators do not register the improvement in bidding. \textit{Payoff Tree} is the second bids-only lever ($3.20 \to 1.29$, $p = 0.017$, all four families) with almost unchanged rule vocabulary ($33.0\% \to 32.7\%$). The \emph{worst-case scaffold}---a contingent-reasoning sibling that instructs the agent to reason about the worst case for each bid rather than laying the cases out---does the opposite: it substantially increases the targeted language (worst-case language $2\% \to 43\%$, $p = 0.0075$) while bids do not move ($3.24 \to 3.07$, $p = 0.81$; equivalent to no change within $\pm\$1.00$), a pooled null that conceals offsetting family effects (Claude and Gemma improve, Gemini and GPT-4o worsen). Within the treated group, plans that do articulate the worst case err no less than plans that do not. The \emph{menu restatement} has no detectable bid effect ($3.55$ vs.\ $3.20$, $p = 0.66$) and its language response is not interpretable as comprehension, since the menu wording removes the ``second-highest'' vocabulary from the rules text and its language column is a prompt echo by construction. The ``both improve'' cell of the primary contrasts is empty.

\paragraph{Exploratory interventions.}
After correcting for multiple tests, the interventions that change both language and bids worsen bidding. \textit{First-Order Beliefs} raises opponent-modeling language from $27.8\%$ to $42.7\%$ of plans ($q = 0.015$) \emph{and} worsens $|b-v|$ by $\$0.76$ ($q = 0.027$), in the same direction in all four families---Gemini included, whose signed mean in Appendix~\ref{sec:beliefs} looked flat; the language rise is partly the prompt's own vocabulary read back. A risk-averse persona does the same (risk language $+70$ points; $|b-v|$ $+\$1.51$). \textit{Second-Order Beliefs} raises opponent language further ($51\%$, $q = 0.020$), but its bid effect does not survive the correction on the absolute metric ($q = 0.36$), making it a words-only lever alongside the risk-seeking persona, which increases aggressive language ($+60$ points, $q = 0.023$) while bids stay flat ($q = 0.93$). One additional bids-only improver appears (the common-knowledge belief prompt, $q = 0.027$, soft under the permutation test) and one previously suggestive backfire does not survive (the endowment frame, $q = 0.10$). Under the same correction across all twenty interventions the two primary improvements attenuate to $q = 0.068$: significant as primary contrasts, but not at the $5\%$ threshold when treated as part of the full exploratory comparison. We report both statements.

\paragraph{Comparing the two outcomes.}
Language predicts the direction of an individual bid once a plan is stated, but treatment-induced changes in language do not track treatment-induced changes in bids. In these comparisons, improved bids are not accompanied by increases in the targeted verbal indicators, increased use of targeted language can occur without better bids, and interventions with detectable changes in both outcomes worsen bidding. For the measured verbal indicators, a classical mediation decomposition estimates the share of \textit{Payoff Safety}'s bid effect that could run through the measured indicators of stated comprehension with a point estimate near zero and a 95\% upper bound of $35\%$, and a corresponding upper bound of $15\%$ for the \textit{Payoff Tree}. These bounds are conditional on the decomposition and its assumptions; they do not bound changes in unobserved understanding. The within-treatment association that plans rehearsing the payment rule err about \$1.4 less is cross-sectional, not causal.

\begin{figure}[htbp]
    \revfloat
    \centering
    \includegraphics[width=\textwidth]{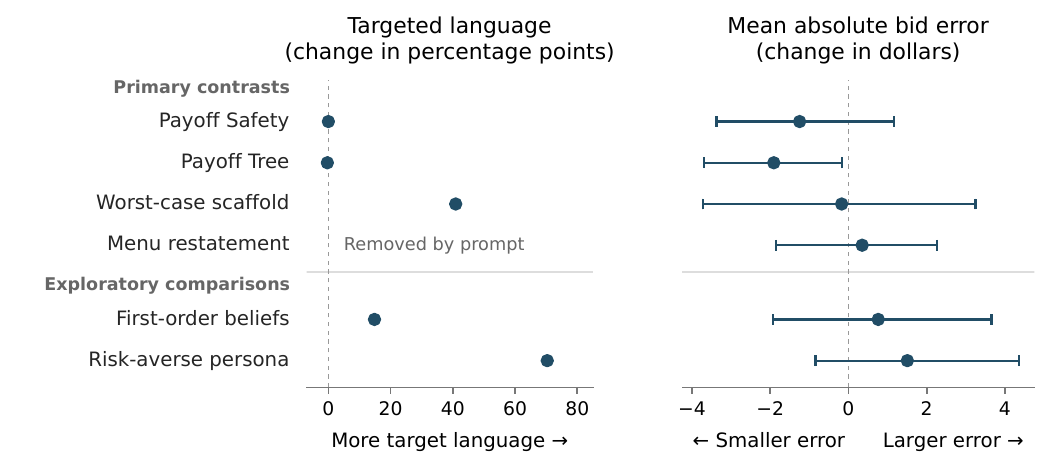}
    \caption{Language and bid responses to six selected interventions (four model families, sealed second-price conditions). Effects are treatment minus baseline. Left: changes in targeted language, shown as point estimates; more target language need not mean better understanding. Right: changes in mean absolute bid error with percentile run-cluster-bootstrap 95\% intervals; negative values indicate smaller errors. The menu removes the measured rule vocabulary, so its language contrast is omitted. The intervals use a different procedure from the wild-bootstrap tests in Table~\ref{tab:dissociation}; all twenty comparisons are in Table~\ref{tab:trace-interventions-full}.}
    \Description{Two aligned panels show language changes and bid-error changes for six interventions.}
    \label{fig:language-bids}
\end{figure}

\begin{table}[htbp]
    \revfloat
    \centering
    \footnotesize
    \setlength{\tabcolsep}{2.5pt}
    \resizebox{\linewidth}{!}{%
    \begin{tabular}{llcccclc}
    \toprule
    & & \multicolumn{2}{c}{Stated reasoning} & \multicolumn{2}{c}{Bids, mean $|b-v|$ (\$)} & \\
    \cmidrule(lr){3-4}\cmidrule(lr){5-6}
    Lever & Targeted language & $\Delta$ prevalence & $p$ / $q$ & base $\to$ treated & $p$ / $q$ & Cell \\
    \midrule
    \multicolumn{7}{l}{\emph{Primary contrasts (raw wild-cluster-bootstrap $p$)}} \\
    Payoff Safety & invariant echo & $0\% \to 0\%$ & --- & $3.20 \to 1.95$ & $p = .015$ & bids only \\
    Payoff Tree & rule rehearsal & $33.0\% \to 32.7\%$ & $p = .90$ & $3.20 \to 1.29$ & $p = .017$ & bids only \\
    Worst-case scaffold & worst-case rhetoric & $2.0\% \to 43.0\%$ & $p = .0075$ & $3.24 \to 3.07$ & $p = .81$ & words only \\
    Menu restatement & rule rehearsal & (by construction) & --- & $3.20 \to 3.55$ & $p = .66$ & neither \\
    \midrule
    \multicolumn{7}{l}{\emph{Exploratory rows (Benjamini--Hochberg $q$, two families)}} \\
    First-order beliefs & opponent modeling & $27.8\% \to 42.7\%$ & $q = .015$ & $3.16 \to 3.93$ & $q = .027$ & both, harmful \\
    Second-order beliefs & opponent modeling & $27.8\% \to 51.3\%$ & $q = .020$ & $3.16 \to 3.87$ & $q = .36$ & words only \\
    Risk-averse persona & risk language & $29.0\% \to 99.3\%$ & $q = .015$ & $3.20 \to 4.71$ & $q = .027$ & both, harmful \\
    Risk-seeking persona & aggressive language & $5.8\% \to 65.5\%$ & $q = .023$ & $3.20 \to 3.49$ & $q = .93$ & words only \\
    \bottomrule
    \end{tabular}}
    \caption{Joint response of stated reasoning and bids for the headline levers (pooled over the four families, sealed second-price cells; run-level wild-cluster bootstrap; the full twenty-row table with per-family sign vectors is Table~\ref{tab:trace-interventions-full} in Appendix~\ref{sec:appendix-traces}). Baselines are the pooled contingent-reasoning and belief baseline cells ($3.20$); the worst-case scaffold is compared with its own grid's baseline ($3.24$) and the belief rows with theirs ($3.16$). Under correction across all twenty interventions the two primary improvements carry $q = 0.068$. No lever occupies the ``both improve'' cell.}
    \label{tab:dissociation}
\end{table}

\end{revblock}

\section{Discussion and Conclusion} \label{sec:conclusion}

\begin{revblock}
Can scaffolds and descriptions improve LLM decision making? Our experiments show that the answer depends on how the support is designed. Human-motivated theories of simplicity identify useful changes to the choices an agent faces and to the rules it is shown. The ascending auction interface improves bidding across four model families; explicit descriptions of payoff contingencies or safety properties improve choices in both auctions and matching. Additional prompts to plan through matching rounds or reason about opponents instead worsen play overall. Effective support therefore requires attention to the structure of the task, rather than an assumption that prompting more reasoning will help.

The auction analysis adds a separate lesson for evaluation. The interventions that improve choices need not improve the measured verbal indicators of strategic understanding in short stated plans. Other interventions change these indicators without improving bids. These findings do not show that the models' internal reasoning stayed fixed: the analysis observes explanations and decisions, not the computations between them. They do show that an explanation alone is insufficient to determine whether an interface change improved behavior. Designers should evaluate realized decisions against an explicit task benchmark and use verbal responses as an additional outcome.

The economic settings make this comparison possible because the rules and the dominant strategies are known. Their broader value is as a testbed for a design method: identify which comparisons a task requires, make relevant consequences accessible at the point of choice, and measure whether the proposed support helps. Our findings suggest that human-motivated theories of simplicity can inform the design of decision environments for artificial agents. Changes to mechanism interfaces and descriptions can improve LLM choices, yet these behavioral gains need not be accompanied by corresponding improvements in measured verbal indicators of strategic understanding.

The scope of this evidence is limited to the tested models, prompts, and environments. Extending the approach to tasks without dominant strategies will require suitable performance benchmarks and may make planning or opponent beliefs useful rather than distracting. The additional model panel in Appendix~\ref{sec:traces-implications} also shows that an intervention's value depends on baseline capability: models already near truthful play have little room to improve, while a poorly interpreted description can still cause errors. The matching interface comparison also requires care: incomplete preference reports and complete rankings do not share an error denominator, and the zero recorded ranking statistic misses some query errors. Within the current testbed, scoring the matching explanations and validating the text labels against a human-labeled sample would test how far the auction findings extend. Combining complementary scaffolds and studying longer interactions would further clarify when support improves decisions and when it introduces new mistakes.
\end{revblock}

\paragraph{Disclosure.} Claude (Anthropic, Opus 4.6) assisted with plotting code and manuscript formatting.

\bibliographystyle{plainnat}
\bibliography{es_bib}

\appendix

\include{contents/appendix}
\end{document}

%% file: contents/appendix.tex

\section{Prompts}
\label{sec:appendix-prompts}

This appendix documents the prompt templates used in our experiments. We present the full base rule prompts for both Deferred Acceptance (DA) and Second-Price Sealed-Bid (SPSB) auction mechanisms, together with the Obviously Strategy-Proof (OSP) sequential interfaces: the \citet{ashlagi2018stable} decision-tree implementation for DA and the \citet{breitmoser2022obviousness} ascending-clock (closed) implementation. 
For each intervention, we show only the text that is \emph{added to or modified from} the corresponding base prompt, since the surrounding context (preference ordering, priorities, output format) remains identical. Template variables (e.g., \texttt{\{\{student\_id\}\}}, \texttt{\{\{preference\_order\}\}}) are populated at runtime with experiment-specific values.

\subsection*{A.1\quad Deferred Acceptance (DA)}

\paragraph{Prompt skeletons.} \rev{The two skeletons below summarize the direct-revelation and iterative prompts whose full templates follow.}
\noindent\textbf{Prompt skeleton (Direct Revelation).}
\begin{quote}\small
You are Student A. There are 4 schools: w, x, y, z. You will be matched to at most one school.

HOW THE MATCHING WORKS (Deferred Acceptance):
(1) All students submit a ranking of schools.
(2) Each student proposes to their top-ranked school.
(3) Each school tentatively accepts the proposer it ranks highest (by priority), rejecting others.
(4) Rejected students propose to their next choice; schools keep the highest-priority student so far.
(5) The algorithm ends when no rejections occur.

Your priority rank (1=highest): \; w:1,\; x:2,\; y:1,\; z:2.\\
Most applicants seem to favor: \; y \textgreater x \textgreater w \textgreater z.\\
(TREATMENT INSTRUCTIONS)

Return:
\texttt{<REASON> ... </\allowbreak REASON>}

\texttt{<DECISION> Ranking: ... </\allowbreak DECISION>}
\end{quote}

\noindent\textbf{Prompt skeleton (Iterative DA, yes/no query).}
\begin{quote}\small
You are Student A. The remaining schools are: w, y, z.

Your preference ordering (most to least preferred): y, w, z.\\
Your priority rank at each school (1=highest): w:1, y:1, z:2.\\
Most applicants seem to favor: y \textgreater{} x \textgreater{} w \textgreater{} z.

Among the remaining schools \{w, y, z\}, is \textbf{y} your most preferred?\\
If YES: you will be matched to y immediately.\\
If NO: y remains available and you will be asked again.

Return:
\texttt{<REASON> ... </\allowbreak REASON>}

\texttt{<DECISION> Answer: YES /\allowbreak  NO </\allowbreak DECISION>}
\end{quote}

\subsubsection*{A.1.1\quad Base Rules --- Direct Revelation}

In the direct revelation interface, the LLM submits a complete ranking of all schools in a single shot. We vary \emph{how} the mechanism is explained while keeping the underlying student-proposing DA algorithm identical.

\paragraph{Null Baseline.}
The minimal prompt provides no explanation of the matching algorithm---only the decision environment and output format.

\begin{figure}[H]
\begin{lstlisting}
You are Student {{student_id}}. There are 4 schools: w, x, y, z.
You will be matched to at most one school based on submitted
rankings and school priorities.

Your preference ordering (most preferred to least preferred):
  {{preference_order}}

Your priority rank at each school (1 = highest priority):
  w: {{pw}}, x: {{px}}, y: {{py}}, z: {{pz}}

Moreover, most applicants seem to favor {{global_ranking}}.

Submit a complete ranking of schools from most preferred to
least preferred.
Format: Ranking: <1st> > <2nd> > <3rd> > <4th>
\end{lstlisting}
\caption{DA Direct Revelation --- Null Baseline (\texttt{da\_\allowbreak direct\_\allowbreak null}). No mechanism explanation is provided.}
\label{fig:da-direct-null}
\end{figure}

\paragraph{Traditional DA.}
The standard prompt describes the full five-step Deferred Acceptance algorithm.

\begin{figure}[H]
\begin{lstlisting}
You are Student {{student_id}}. There are 4 schools: w, x, y, z.
You will be matched to at most one school.

HOW THE MATCHING WORKS (Deferred Acceptance):
1. All students simultaneously submit a ranking of schools
2. Each student "proposes" to their top-ranked school
3. Each school tentatively accepts the proposer it ranks highest
   (by priority), rejecting others
4. Rejected students propose to their next choice
5. This repeats until all students are matched

Your preference ordering (most preferred to least preferred):
  {{preference_order}}

Your priority rank at each school (1 = highest priority):
  w: {{pw}}, x: {{px}}, y: {{py}}, z: {{pz}}

Moreover, most applicants seem to favor {{global_ranking}}.
\end{lstlisting}
\caption{DA Direct Revelation --- Traditional (\texttt{da\_\allowbreak direct\_\allowbreak traditional}). The full student-proposing DA algorithm is described step by step.}
\label{fig:da-direct-traditional}
\end{figure}

\subsubsection*{A.1.2\quad OSP Sequential Interface for DA}

Our OSP implementation for DA follows the decision-tree construction of \citet{ashlagi2018stable}, which provides an OSP mechanism for student-proposing DA under Ergin-acyclic priority profiles. Rather than asking the LLM to submit a full ranking, the mechanism queries \emph{one student at a time} with binary yes/no offers about individual schools, each accompanied by an explicit guarantee of what happens under either response. Algorithm~\ref{alg:osp-da} summarizes the procedure.

\begin{algorithm}[H]
\caption{OSP Decision Tree for DA under Acyclic Priorities \citep{ashlagi2018stable}}
\label{alg:osp-da}
\begin{algorithmic}[1]
\Procedure{RunOSPTree}{students, schools, priorities}
  \If{students $= \emptyset$}
    \State \Return matches
  \EndIf
  \State $\text{top\_by\_school} \gets \{s : \arg\min_{\text{priority}} \text{students at } s\}$ for each remaining school $s$
  \State $\text{top\_students} \gets \text{unique}(\text{top\_by\_school.values})$
  \If{$|\text{top\_students}| = 1$} \Comment{Serial dictatorship node}
    \State $s^* \gets$ the single top-priority student
    \State $w \gets$ \Call{AskPickTop}{$s^*$, schools} \Comment{Use \texttt{da\_\allowbreak osp\_\allowbreak choice} prompt}
    \State Assign $s^*$ to $w$
    \State \Return \Call{RunOSPTree}{students $\setminus \{s^*\}$, schools $\setminus \{w\}$, priorities}
  \EndIf
  \State Let $a, b \gets \text{top\_students}$ \Comment{$|\text{top\_students}| = 2$ under acyclicity}
  \For{each school $w$ where $\text{top\_by\_school}[w] = a$ (in fixed order)}
    \If{\Call{AskYesNo}{$a$, $w$, fallback $=$ schools}} \Comment{Use \texttt{da\_\allowbreak osp\_\allowbreak yesno\_\allowbreak guaranteed}}
      \State Assign $a$ to $w$
      \State \Return \Call{RunOSPTree}{students $\setminus \{a\}$, schools $\setminus \{w\}$, priorities}
    \EndIf
  \EndFor
  \For{each school $w$ where $\text{top\_by\_school}[w] = b$ (in fixed order)}
    \If{\Call{AskYesNo}{$b$, $w$, fallback $=$ schools}}
      \State Assign $b$ to $w$
      \State \Return \Call{RunOSPTree}{students $\setminus \{b\}$, schools $\setminus \{w\}$, priorities}
    \EndIf
  \EndFor
  \State $w_a \gets$ \Call{AskPickTop}{$a$, schools}; assign $a$ to $w_a$
  \State $w_b \gets$ \Call{AskPickTop}{$b$, schools $\setminus \{w_a\}$}; assign $b$ to $w_b$
  \State \Return \Call{RunOSPTree}{students $\setminus \{a,b\}$, schools $\setminus \{w_a, w_b\}$, priorities}
\EndProcedure
\end{algorithmic}
\end{algorithm}

\rev{The relevant guarantee concerns the quality of the eventual assignment. YES immediately assigns the offered school. If that school is not the student's favorite, truthful continuation after NO guarantees an outcome at least as preferred as accepting the offer, under the construction of \citet{ashlagi2018stable}. NO does not immediately remove the school, but does not reserve every school indefinitely: another student may be assigned a school before the next query. The historical prompt below uses the shorthand ``remains available''; we reproduce it as used rather than treating that wording as the formal guarantee.}

\paragraph{Yes/No with Consequence Explanation.}
The primary prompt used at binary-offer nodes in Algorithm~\ref{alg:osp-da}. It explicitly states the consequence of each answer, providing the ``obviousness witness.''

\begin{figure}[H]
\begin{lstlisting}
You are Student {{student_id}}. We are running a step-by-step
matching procedure.

Current remaining schools: {{remaining_set}}

Your preference ordering (most preferred to least preferred):
  {{preference_order}}

Your priority rank at each school (1 = highest priority):
  w: {{pw}}, x: {{px}}, y: {{py}}, z: {{pz}}

Moreover, most applicants seem to favor {{global_ranking}}.

Question: Among the remaining schools ({{remaining_set}}), is
  {{candidate}} your most preferred?
- If you answer YES: you are immediately matched to
  {{candidate}} and leave the process.
- If you answer NO: {{candidate}} remains available to you.
  You continue in the process with all current remaining
  schools ({{fallback_set}}).

Please provide your response in TWO parts:

1. First, explain your reasoning in <REASON></REASON> tags
2. Then, submit your answer in <DECISION></DECISION> tags

Format:
<REASON>
Your reasoning here...
</REASON>

<DECISION>
Answer: YES
</DECISION>

or

<DECISION>
Answer: NO
</DECISION>
\end{lstlisting}
\caption{DA OSP --- Yes/No with Consequences (\texttt{da\_\allowbreak osp\_\allowbreak yesno}). Historical prompt describing YES and NO consequences. The blanket availability wording is stronger than the formal continuation guarantee; see the clarification above.}
\label{fig:da-osp-yesno-guaranteed}
\end{figure}

\subsubsection*{A.1.3\quad Reasoning Interventions for DA}

Each intervention below is appended to (or replaces a section of) the Traditional DA base prompt (Figure~\ref{fig:da-direct-traditional}). The surrounding context---student identity, preference ordering, priorities, global ranking hint, and output format---remains identical. We show only the intervention-specific text that differs from the base.

\paragraph{Axis 1: Contingent Reasoning.}
These interventions prompt the agent to reason about what happens given others' possible actions. We show the \emph{Enumerate} variant; other Axis~1 variants are listed in Table~\ref{tab:all-interventions}.

\begin{figure}[H]
\begin{lstlisting}
REASONING GUIDANCE:
Before submitting your ranking, consider: What rankings might
the other students submit? For each possible combination of
their rankings, what school would you receive if you ranked
truthfully vs. if you ranked differently?
\end{lstlisting}
\caption{DA Axis 1 --- Enumerate (\texttt{axis1\_\allowbreak da\_\allowbreak enumerate}). Inserted after the DA algorithm description. Other variants: \emph{Dominated} (eliminate dominated strategies), \emph{Worst-Case} (consider minimum obtainable set), \emph{One-Step} (simplify to single-step decision), \emph{Decision Tree} (visualize DA as a tree), \emph{Backward Induction} (reason backwards from final round).}
\label{fig:da-axis1-enumerate}
\end{figure}

\paragraph{Axis 2: Forward Planning.}
These interventions scaffold varying depths of forward simulation through the DA algorithm. We show the $k{=}1$ variant; other Axis~2 variants are listed in Table~\ref{tab:all-interventions}.

\begin{figure}[H]
\begin{lstlisting}
HOW THE MATCHING WORKS:
You submit a ranking. The algorithm processes your ranking
sequentially:
- First, you "propose" to your top-ranked school
- If accepted, you're matched there
- If rejected, you propose to your second choice

THINK ONE STEP AHEAD:
Before finalizing your ranking, consider: If you are rejected
from your first choice, which school would you propose to next?
Does this affect how you should rank schools?
\end{lstlisting}
\caption{DA Axis 2 --- $k{=}1$ Forward Planning (\texttt{axis2\_\allowbreak da\_\allowbreak 1step}). Replaces the standard DA algorithm description with a simplified sequential framing plus one-step lookahead guidance. Other variants: $k{=}0$ (no guidance), $k{=}2$ (two-step lookahead), $k{=}\infty$ (full simulation), and three \emph{Monotonicity} framings (options never shrink; rejections redirect, never eliminate; outcome determined by priorities).}
\label{fig:da-axis2-1step}
\end{figure}

\paragraph{Axis 3: Higher-Order Beliefs.}
These interventions prompt reasoning about other agents' beliefs and common knowledge. We show the \emph{Common Knowledge} variant.

\begin{figure}[H]
\begin{lstlisting}
IMPORTANT: All students know these rules. All students know
that all students know these rules. All students are trying
to maximize their own earnings.
\end{lstlisting}
\caption{DA Axis 3 --- Common Knowledge (\texttt{axis3\_\allowbreak da\_\allowbreak common\_\allowbreak knowledge}). Inserted after the DA algorithm description. Other variants: \emph{First-Order} (what will others rank?) and \emph{Second-Order} (what do others believe you will rank?).}
\label{fig:da-axis3-common-knowledge}
\end{figure}

\newpage
\subsection*{A.2\quad Second-Price Sealed-Bid Auction}

\paragraph{Prompt skeletons.} \rev{The two skeletons below summarize the sealed-bid and clock prompts whose full templates follow.}
\noindent\textbf{Prompt skeleton (SPSB).}
\begin{quote}\small
You are Bidder \{Name\}. You are bidding against two other bidders.\\
MECHANISM DESCRIPTION + TREATMENT TEXT\\
Your private value for the item is \${$v_i$}.\\
Return:
\texttt{<PLAN> ... </\allowbreak PLAN>}\\
\texttt{<ACTION> [your bid] </\allowbreak ACTION>}
\end{quote}

\noindent\textbf{Prompt skeleton (Clock, at posted price $p$).}
\begin{quote}\small
You are Bidder \{Name\}. The current posted price is \${$p$}.\\
If you stay, you remain in the auction at this price. If you exit, you leave permanently.
You will not be told when other bidders exit.\\
MECHANISM DESCRIPTION + TREATMENT TEXT\\
Your private value for the item is \${$v_i$}.\\
Return:
\texttt{<PLAN> ... </\allowbreak PLAN>}\\
\texttt{<ACTION> Stay /\allowbreak  Exit </\allowbreak ACTION>}
\end{quote}

\subsubsection*{A.2.1\quad Base Rule}
The base rule is implemented as the standard Second-price sealed-bid auction.

\begin{figure}[H]
\begin{lstlisting}
In this game, you will participate in an auction for a prize
against {{num_bidders}} other bidders. You will play this game
for {{n}} rounds.
At the start of each round, bidders will see their value for
the prize, randomly drawn between $0 and ${{private}}, with
all values equally likely.
After learning your value, you will submit a bid privately at
the same time as the other bidders. Bids must be in
${{increment}} increments.
The highest bidder wins the prize and pays the second-highest
bid. If you win, your earnings will increase by your value for
the prize, and decrease by the second-highest bid. If you don't
win, your earnings will remain unchanged.
After each auction, we will display all bids. Ties for the
highest bid will be resolved randomly.
\end{lstlisting}
\caption{SPSB Auction --- Base Rule (\texttt{private\_\allowbreak second\_\allowbreak price}). Standard second-price sealed-bid auction with independent private values.}
\label{fig:auction-base}
\end{figure}

\subsubsection*{A.2.2\quad OSP Implementation --- Two-Stage Ascending Clock}

Following \citet{breitmoser2022obviousness}, we implement an Ascending Clock Auctions, each round consists of multiple clock cycles, during which every bidder is asked whether they want to stay or drop out at the current clock price.
In the first round, bidders are reminded of the auction rules. 
The auction starts with an initial price of 0, which increases incrementally until only one bidder remains or two bidders drop out simultaneously, in which case the winner is chosen randomly. 
The detailed prompt is listed as follows, with variables enclosed in brackets:

\begin{figure}[H]
\begin{lstlisting}
In this game, you will participate in an auction for a prize
against {{num_bidders}} other bidders. You will play this game
for {{n}} rounds.
At the start of each round, bidders will see their value for
the prize, randomly drawn between $0 and ${{private}}, with
all values equally likely.

**END OF AUCTION**
If you win, your earnings will increase by your value for the
prize and decrease by the clock price at the end of the auction.
If you don't win, your earnings will remain unchanged.
After each auction, we will display all bids. Ties for the
highest bid will be resolved randomly.

Your value towards to the prize is {value} in this round.
The current price in this clock cycle is {current_price}. 
The price for next clock cycle is {current_price + increment}.

Do you want to stay in the bidding? 
If you choose yes, you can keep bidding for next clock. If you choose No, you will exit and have no chance to re-enter the bidding. Your response must use these EXACT tags below. You must output the ACTION.
    ```
    <PLAN>
    [Write your plans for bidding strategies. Be detailed and precise but keep things succinct and don't repeat yourself. LIMIT your plan to 50 words. ] </PLAN>
    <ACTION> Yes or No </ACTION>
\end{lstlisting}
\caption{SPSB OSP --- Ascending Clock (\texttt{intervention\_\allowbreak proxy\_\allowbreak breitmoser}). No dropping out information will be displayed as in \citet{breitmoser2022obviousness}.}
\label{fig:auction-proxy}
\end{figure}

\subsubsection*{A.2.3\quad Reasoning Interventions for SPSB} \label{sec:appendix-prompts-auction}

Each intervention below is appended to or modifies the SPSB base prompt (Figure~\ref{fig:auction-base}). We show only the intervention-specific text.

\paragraph{Axis 1: Contingent Reasoning.}
We show the \emph{Enumerate} variant. Other Axis~1 variants are listed in Table~\ref{tab:all-interventions}.

\begin{figure}[H]
\begin{lstlisting}
Before deciding your bid, think through the following: What are
the possible bids the other players might submit? For each
possible bid they might make, what would be your best response?
Does your optimal bid depend on what others do, or is there a
strategy that works well regardless?
\end{lstlisting}
\caption{SPSB Axis 1 --- Enumerate (\texttt{axis1\_\allowbreak contingent\_\allowbreak enumerate}). Inserted before the ``After each auction'' line. Other variants: \emph{Dominated} (identify dominated bids), \emph{Worst-Case} (worst-case analysis per bid), \emph{One-Step} (bid as ``maximum price''), \emph{Decision Tree} (PATH A: win vs.\ PATH B: lose), \emph{Backward Induction} (two-stage sealed $\to$ clock with backward-induction guidance).}
\label{fig:auction-axis1-enumerate}
\end{figure}

\paragraph{Axis 3: Higher-Order Beliefs.}
We show the \emph{Common Knowledge} variant.

\begin{figure}[H]
\begin{lstlisting}
Important: All bidders are rational and this is common
knowledge. This means:
- Every bidder is rational and maximizes their own earnings
- Every bidder knows that every other bidder is rational
- Every bidder knows that every other bidder knows this
- And so on, infinitely

Given this common knowledge of rationality, what is the optimal
bidding strategy? Is there a strategy that all rational bidders
would converge on?
\end{lstlisting}
\caption{SPSB Axis 3 --- Common Knowledge (\texttt{axis3\_\allowbreak beliefs\_\allowbreak common\_\allowbreak knowledge}). Inserted after the payment rule. Other variants: \emph{First-Order} (what do others bid?) and \emph{Second-Order} (what do others think you bid?).}
\label{fig:auction-axis3-common-knowledge}
\end{figure}

\begin{revblock}
\paragraph{Mechanism Description and Worst-Case Scaffold (trace analysis).}
Two further prompts enter the set of interventions of Section~\ref{sec:traces}. The \emph{menu restatement} replaces the payment paragraph of the base rule (Figure~\ref{fig:auction-base}) with a menu description in the spirit of \citet{gonczarowski2023strategyproofness}; the \emph{worst-case} contingent-reasoning scaffold is the \texttt{axis1\_\allowbreak contingent\_\allowbreak worstcase} variant listed in Table~\ref{tab:all-interventions}, inserted before the ``After each auction'' line.

\begin{figure}[H]
\revfloat
\begin{lstlisting}
Your "price to win" the item will be set to the highest bid
placed by any other player. If your bid is higher than this
"price to win," then you will win the item and pay this price.
If you don't win, your earnings will remain unchanged.
\end{lstlisting}
\caption{SPSB Mechanism Description --- Menu Restatement (\texttt{intervention\_\allowbreak menu}). Replaces the payment paragraph of the base rule; note that it removes the words ``second-highest'' from the rules text, which is why its language column in the trace analysis is uninterpretable by construction (Section~\ref{sec:traces-probe}).}
\label{fig:auction-menu}
\end{figure}

\begin{figure}[H]
\revfloat
\begin{lstlisting}
Before deciding your bid, consider the worst-case scenario for
each possible bid you could make. What is the worst that could
happen if you bid X? Find the bid where even the worst-case
outcome is acceptable. Compare: if you deviate from this bid,
could the worst-case be worse?
\end{lstlisting}
\caption{SPSB Axis 1 --- Worst-Case (\texttt{axis1\_\allowbreak contingent\_\allowbreak worstcase}). A contingent-reasoning scaffold that asks the bidder to construct the worst case for each bid rather than laying the cases out; it passes its manipulation check on stated plans but leaves bids unchanged (Section~\ref{sec:traces-probe}).}
\label{fig:auction-axis1-worstcase}
\end{figure}
\end{revblock}

\subsection*{A.3\quad Summary of All Prompt Variants}

Table~\ref{tab:all-interventions} lists every prompt template used in the experiments. Prompts shown in full above are marked with $\star$; prompts whose intervention-specific text is shown are marked with $\dagger$.

\setlength{\tabcolsep}{4pt}
\begin{longtable}{p{2.9cm} p{4.6cm} p{5.2cm}}
\toprule
\textbf{Category} & \textbf{Prompt ID} & \textbf{Description} \\
\midrule
\endfirsthead
\toprule
\textbf{Category} & \textbf{Prompt ID} & \textbf{Description} \\
\midrule
\endhead
\bottomrule
\endlastfoot

\multicolumn{3}{l}{\textbf{Deferred Acceptance --- Direct Revelation}} \\
\midrule
Base & da\_\allowbreak direct\_\allowbreak null $\star$ & Minimal baseline; no mechanism explanation \\
Base & da\_\allowbreak direct\_\allowbreak traditional $\star$ & Full 5-step DA algorithm description \\
\midrule

\multicolumn{3}{l}{\textbf{Deferred Acceptance --- OSP Sequential \citep{ashlagi2018stable}}} \\
\midrule
OSP & da\_\allowbreak osp\_\allowbreak yesno $\star$ & Binary with consequence explanation (obviousness witness) \\
\midrule

\multicolumn{3}{l}{\textbf{DA Axis 1: Contingent Reasoning}} \\
\midrule
Axis 1 & axis1\_\allowbreak da\_\allowbreak enumerate $\dagger$ & Enumerate others' possible rankings \\
Axis 1 & axis1\_\allowbreak da\_\allowbreak dominated & Identify and eliminate dominated rankings \\
Axis 1 & axis1\_\allowbreak da\_\allowbreak worstcase & Consider worst-case obtainable set \\
Axis 1 & axis1\_\allowbreak da\_\allowbreak onestep & Simplify to one-step decision \\
Axis 1 & axis1\_\allowbreak da\_\allowbreak tree & Visualize DA as a decision tree \\
Axis 1 & axis1\_\allowbreak da\_\allowbreak backward\_\allowbreak induct & Reason backwards from final round \\
\midrule

\multicolumn{3}{l}{\textbf{DA Axis 2: Forward Planning}} \\
\midrule
Axis 2 & axis2\_\allowbreak da\_\allowbreak 0step & $k{=}0$: no guidance baseline \\
Axis 2 & axis2\_\allowbreak da\_\allowbreak 1step $\dagger$ & $k{=}1$: think one step ahead \\
Axis 2 & axis2\_\allowbreak da\_\allowbreak 2step & $k{=}2$: think two steps ahead \\
Axis 2 & axis2\_\allowbreak da\_\allowbreak fullsim & $k{=}\infty$: simulate full algorithm \\
Axis 2 & axis2\_\allowbreak da\_\allowbreak monotonic\_\allowbreak options & ``Your options never shrink'' \\
Axis 2 & axis2\_\allowbreak da\_\allowbreak monotonic\_\allowbreak safety & ``Rejections redirect, never eliminate'' \\
Axis 2 & axis2\_\allowbreak da\_\allowbreak monotonic\_\allowbreak outcome & ``Outcome determined by priorities, not ranking'' \\
\midrule

\multicolumn{3}{l}{\textbf{DA Axis 3: Higher-Order Beliefs}} \\
\midrule
Axis 3 & axis3\_\allowbreak da\_\allowbreak firstorder & What will others rank? \\
Axis 3 & axis3\_\allowbreak da\_\allowbreak secondorder & What do others believe you will rank? \\
Axis 3 & axis3\_\allowbreak da\_\allowbreak common\_\allowbreak knowledge $\dagger$ & Common knowledge of rationality \\
\midrule

\multicolumn{3}{l}{\textbf{SPSB Auction --- Base}} \\
\midrule
Base & private\_\allowbreak second\_\allowbreak price $\star$ & Standard SPSB with independent private values \\
\midrule

\multicolumn{3}{l}{\textbf{SPSB --- OSP Implementations \citep{breitmoser2022obviousness}}} \\
\midrule
OSP & intervention\_\allowbreak proxy\_\allowbreak breitmoser $\star$ & ascending clock \\
\midrule
\multicolumn{3}{l}{\rev{\textbf{SPSB --- Mechanism Descriptions (trace analysis)}}} \\
\midrule
\rev{Description} & \rev{intervention\_\allowbreak menu $\dagger$} & \rev{Menu restatement: ``price to win'' set by the other bids \citep{gonczarowski2023strategyproofness}} \\

\midrule

\multicolumn{3}{l}{\textbf{SPSB Axis 1: Contingent Reasoning}} \\
\midrule
Axis 1 & axis1\_\allowbreak contingent\_\allowbreak baseline & SPSB baseline (identical to base) \\
Axis 1 & axis1\_\allowbreak contingent\_\allowbreak enumerate $\dagger$ & Enumerate others' possible bids \\
Axis 1 & axis1\_\allowbreak contingent\_\allowbreak dominated & Identify and eliminate dominated bids \\
Axis 1 & axis1\_\allowbreak contingent\_\allowbreak worstcase & Consider worst-case for each bid \\
Axis 1 & axis1\_\allowbreak contingent\_\allowbreak onestep & Bid as ``maximum price'' framing \\
Axis 1 & axis1\_\allowbreak contingent\_\allowbreak tree & Decision tree: PATH A (win) vs.\ PATH B (lose) \\
Axis 1 & axis1\_\allowbreak contingent\_\allowbreak backward\_\allowbreak induct & Two-stage sealed $\to$ clock with backward induction \\
\midrule

\multicolumn{3}{l}{\textbf{SPSB Axis 3: Higher-Order Beliefs}} \\
\midrule
Axis 3 & axis3\_\allowbreak beliefs\_\allowbreak baseline & SPSB with ``rational agents'' mention \\
Axis 3 & axis3\_\allowbreak beliefs\_\allowbreak firstorder & What do others bid? \\
Axis 3 & axis3\_\allowbreak beliefs\_\allowbreak secondorder & What do others think you bid? \\
Axis 3 & axis3\_\allowbreak beliefs\_\allowbreak common\_\allowbreak knowledge $\dagger$ & Common knowledge of rationality \\

\bottomrule
\caption{Complete list of prompt templates. $\star$~=~full prompt shown; $\dagger$~=~intervention-specific text shown.}
\label{tab:all-interventions}
\end{longtable}

\section{Prospect-Theoretic Interventions} \label{sec:appendix-prospect}

In addition to the cognitive interventions reported in the main text, we test a separate family of interventions inspired by prospect theory \citep{kahneman1979prospect} and behavioral departures from expected utility theory. These interventions do not scaffold strategic reasoning; instead, they reframe the payoff structure to test whether LLMs exhibit sensitivity to framing effects, loss aversion, the endowment effect, and risk attitudes. We test these across both auctions and DA, with the same four model families.

\subsection*{B.1\quad Loss Aversion and Framing Effects}

We test five variants that reframe the payoff description while holding the mechanism and underlying payoffs constant:

\begin{itemize}
    \item \emph{Gain Frame}: Presents all outcomes as gains from zero (``you cannot lose money---you can only gain'').
    \item \emph{Loss Frame}: Endows the agent with an initial amount equal to their maximum possible value, then presents outcomes as losses from this reference point (``Think carefully about what you might lose'').
    \item \emph{Mixed Frame}: Explicitly decomposes each outcome into a GAIN component and a LOSE component, asking the agent to weigh both.
    \item \emph{Endowment}: Gives the agent an initial endowment and frames payment as a deduction ``from your endowment,'' targeting the endowment effect.
    \item \emph{WTA/WTP}: Compares the agent's willingness to accept a selling price with its willingness to pay a buying price, invoking the WTA--WTP gap.
\end{itemize}

\begin{figure}[H]
    \centering
    \includegraphics[width=\textwidth]{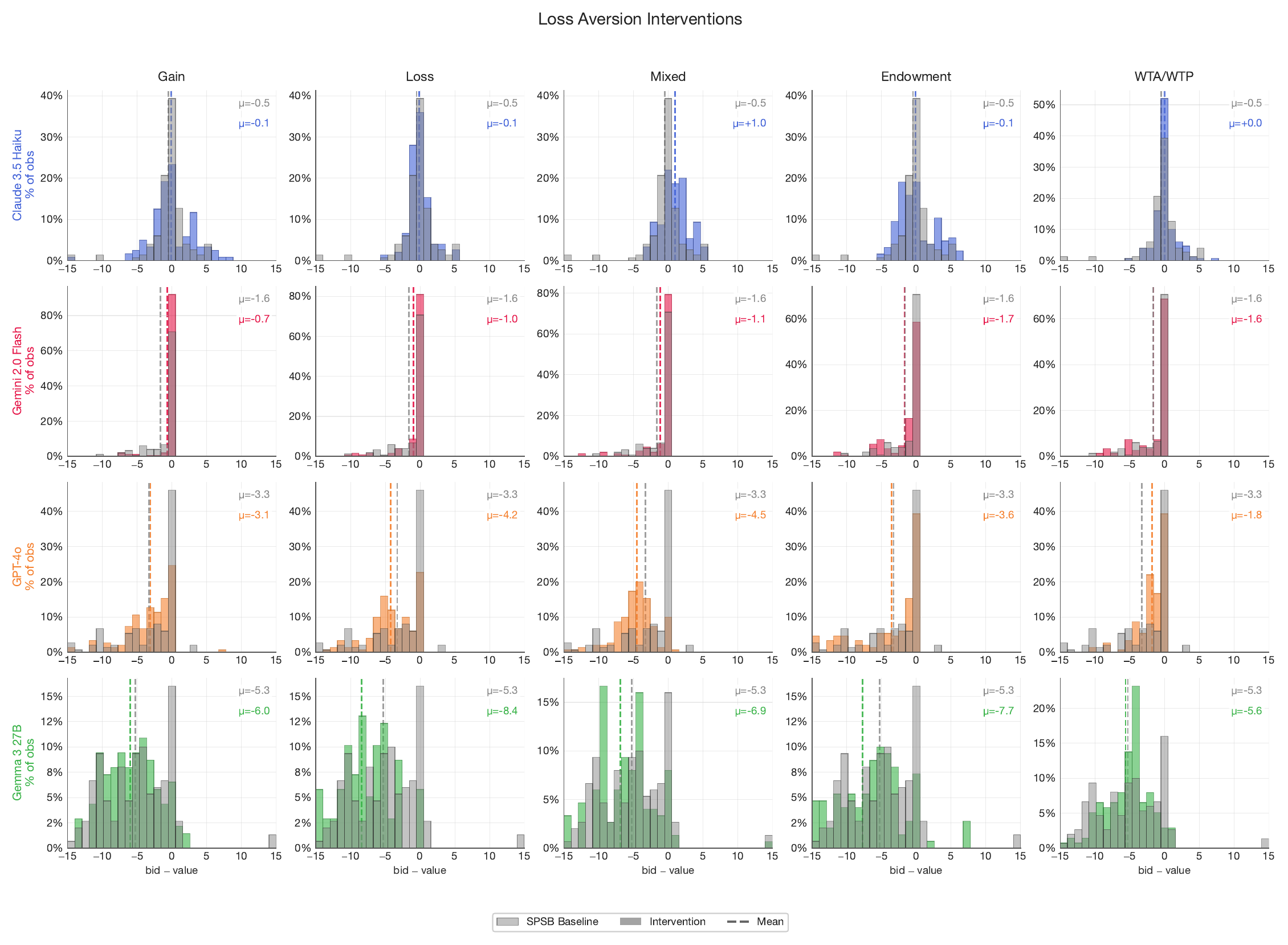}
    \caption{Loss aversion interventions in SPSB auctions. Baseline SPSB in gray, intervention in color. Dashed lines indicate means. None of the framing interventions consistently improve play relative to the SPSB baseline ($\mu = -2.67$). The loss frame ($\mu = -3.44$) and endowment ($\mu = -3.40$) worsen underbidding, while the gain frame ($\mu = -2.51$) and WTA/WTP ($\mu = -2.17$) show modest, inconsistent effects.}
    \label{fig:appendix-loss-auctions}
\end{figure}

\begin{figure}[H]
    \centering
    \includegraphics[width=\textwidth]{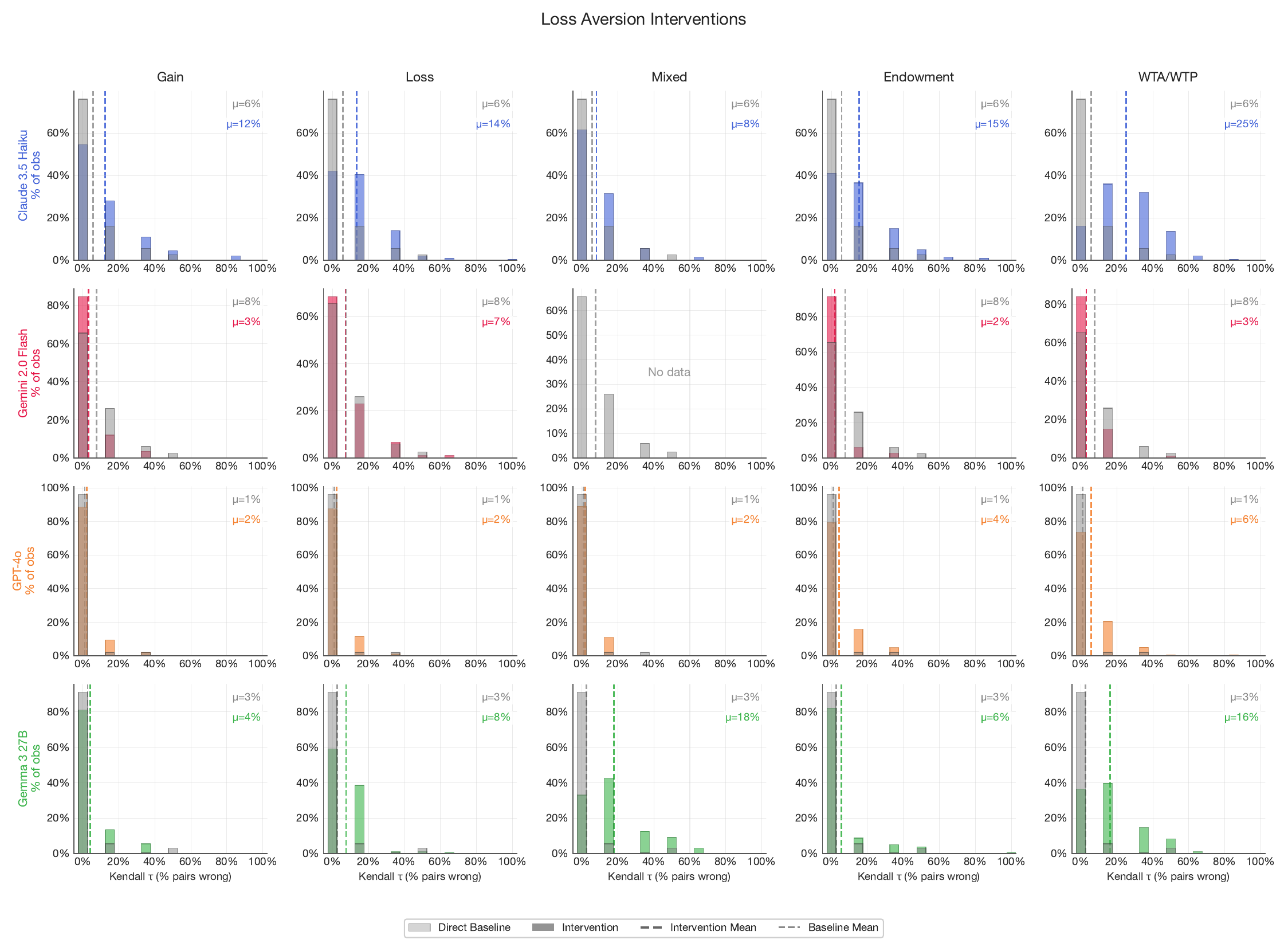}
    \caption{Loss aversion interventions in deferred acceptance. Baseline direct DA in gray, intervention in color. The results are noisy and largely negative: most framing interventions worsen play relative to the baseline ($\mu = 4.2\%$), with WTA/WTP producing the worst outcomes ($\mu \approx 12.6\%$ across models with data). Claude is particularly sensitive to these framings.}
    \label{fig:appendix-loss-da}
\end{figure}

In auctions, the framing interventions do not produce consistent improvements. The loss frame and endowment interventions worsen underbidding (overall means of $\mu = -3.44$ and $-3.40$, compared to the SPSB baseline of $-2.67$), opposite to what a naive application of loss aversion would predict. The gain frame and WTA/WTP interventions produce small, model-dependent effects. In DA, the interventions are uniformly harmful, with WTA/WTP and endowment producing the largest increases in error.

The overall picture is that prospect-theoretic framings do not improve play---and often worsen it. This contrasts with the main-text interventions along contingent reasoning and mechanism description, which produce large, consistent improvements. In these experiments, descriptions of payoff contingencies and incentive properties are more useful than the tested changes in gain--loss framing.

\subsection*{B.2\quad Risk Preference Interventions}

We also test whether assigning explicit risk attitudes to LLMs changes bidding behavior. Each intervention instructs the model to adopt a calibrated risk preference via a concrete coin-toss example:

\begin{itemize}
    \item \emph{Risk Averse}: ``You would only pay \$4 for a coin toss worth \$0 or \$10 (expected value \$5).''
    \item \emph{Risk Neutral}: ``You would pay \$5 for a coin toss worth \$0 or \$10.''
    \item \emph{Risk Seeking}: ``You would pay \$6 for a coin toss worth \$0 or \$10 (expected value \$5).''
\end{itemize}

\begin{figure}[H]
    \centering
    \includegraphics[width=\textwidth]{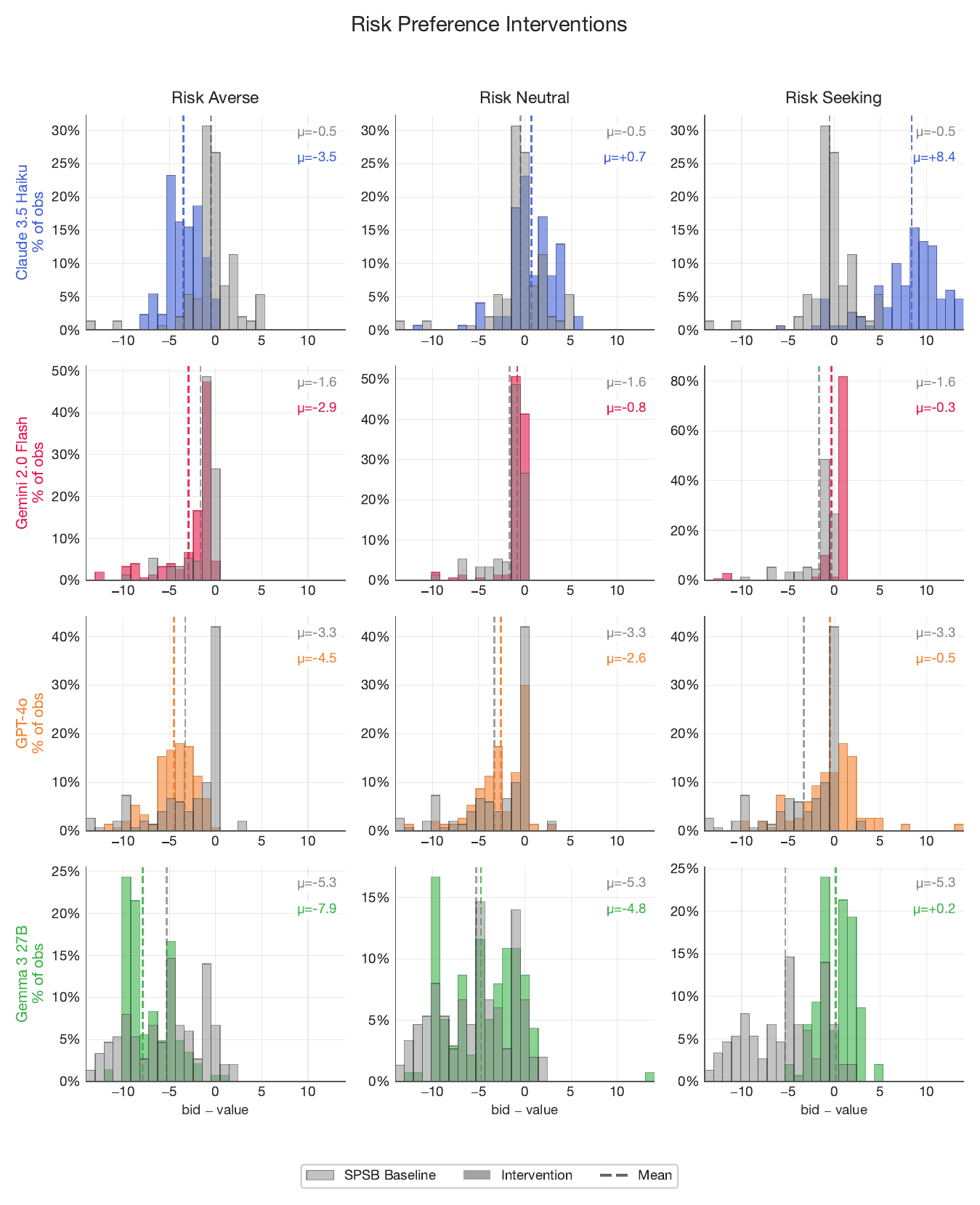}
    \caption{Risk preference interventions in SPSB auctions. Risk aversion increases underbidding ($\mu = -4.71$), while risk-seeking dramatically shifts Claude to overbidding ($\mu = +8.40$). Risk neutrality ($\mu = -1.82$) modestly improves play relative to the SPSB baseline ($\mu = -2.67$).}
    \label{fig:appendix-risk-auctions}
\end{figure}

\begin{figure}[H]
    \centering
    \includegraphics[width=\textwidth]{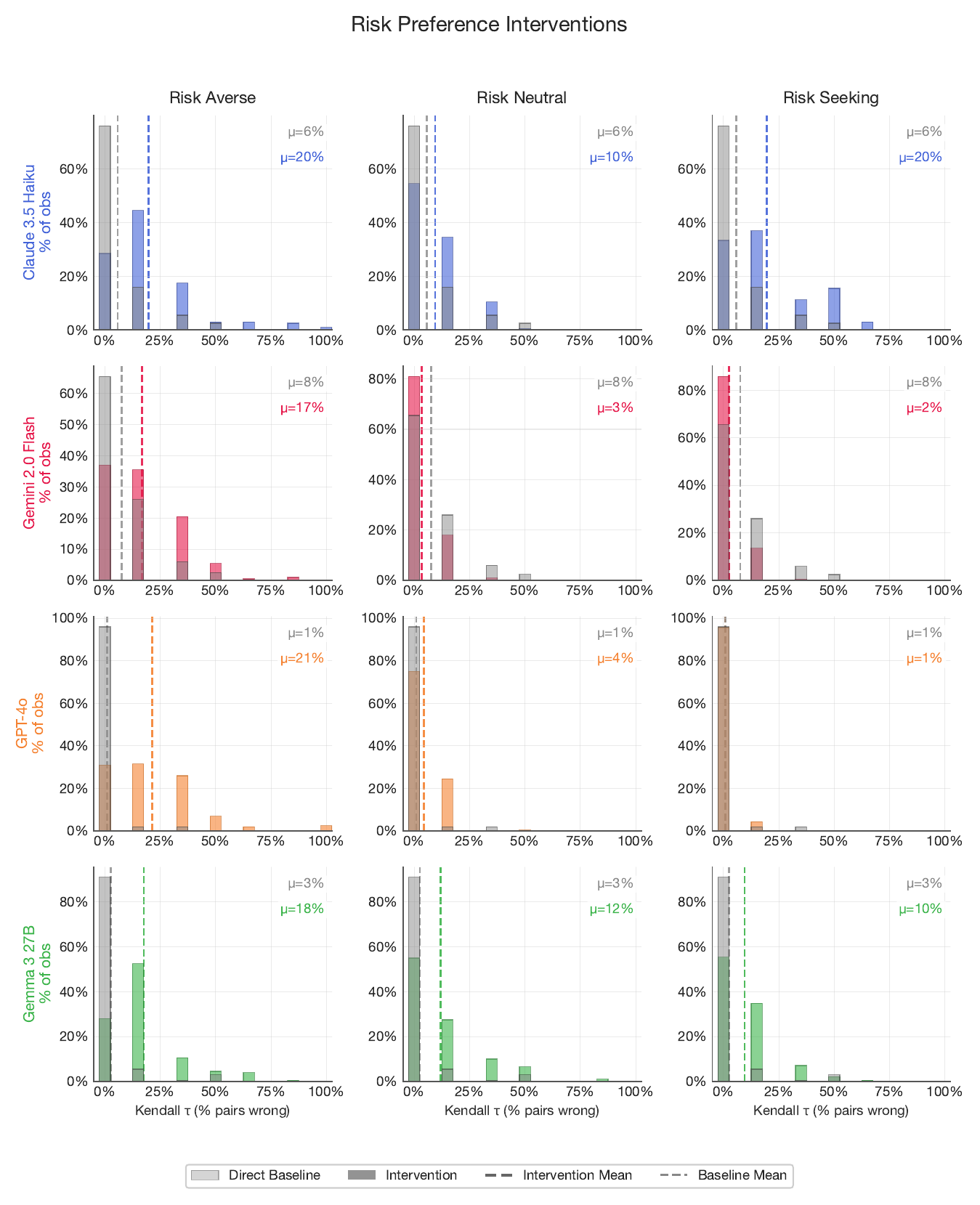}
    \caption{Risk preference interventions in deferred acceptance. Risk aversion dramatically worsens play ($\mu \approx 18.8\%$, compared to baseline $4.2\%$). Risk neutrality and risk-seeking also worsen play, though less severely.}
    \label{fig:appendix-risk-da}
\end{figure}

The risk interventions reveal that LLMs are highly responsive to persona-based instructions about risk attitudes, but that this responsiveness does not improve play. The risk-averse persona produces the most dramatic deterioration: in auctions, it increases underbidding from $\mu = -2.67$ to $-4.71$; in DA, it roughly quadruples the error rate. The risk-seeking persona shifts Claude 3.5 Haiku to massive overbidding ($\mu = +8.40$), while the other models are less affected. Risk neutrality modestly improves auction play ($\mu = -1.82$) but worsens DA play. These comparisons show that instructions about risk attitudes can substantially change behavior. 

Truthful bidding remains dominant for bidders whose utility increases with monetary payoff in this private-value setting. The risk-persona instructions therefore introduce behavioral changes that are not needed to accommodate risk attitudes under the stated rules. These changes do not, by themselves, identify the models' understanding or the cause of their baseline errors.

\begin{revblock}
\section{Reasoning-Trace Analysis: Details} \label{sec:appendix-traces}

This appendix explains how the analysis in Section~\ref{sec:traces} connects stated plans to bids. It describes the corpus, text labels, judge validation, intervention effects, mediation estimates, robustness checks, and additional models. Plans remain verbal outputs, not direct observations of computation.

Replication scripts build the corpus, score text features, classify plans, and estimate intervention effects. Their outputs are stored under \texttt{results/\allowbreak traces/}.

\subsection*{C.1\quad Corpus}

The corpus contains $21{,}990$ plans from three sources: (i)~$13{,}416$ plans from the harmonized four-model grid (92 run directories covering the sealed-bid SPSB baseline and intervention cells plus the closed clock); (ii)~$1{,}167$ plans from the menu-restatement and clock-framing cells, all four models; and (iii)~$7{,}407$ recovered GPT-4o plans from the first- and third-price variants of the same grid, which provide the cross-mechanism control of Section~\ref{sec:traces-scripts}. Duplicate second-price cells present in both (i) and (iii) were excluded from (iii). Each plan carries its run directory as a cluster identifier. The $14{,}073$ sealed second-price plans form the analysis pool for all share and regression statistics; the $510$ ascending-clock rationales are per-decision exit statements and are scored separately (Section~\ref{sec:traces-clock}). The corpus and dictionary were frozen before the intervention analysis (manifest hash in \texttt{results/\allowbreak traces/\allowbreak corpus\_\allowbreak manifest.json}); the frontier extension of $5{,}985$ plans (GPT-5, GPT-5 mini, Claude Sonnet 5, Gemini 2.5 Flash) and an $8{,}616$-plan per-format robustness grid were scored afterwards with the same frozen dictionary plus one additive feature group for formal derivation.

\subsection*{C.2\quad Feature Dictionary, Decision Scripts, and Model Fingerprints}
\applabel{C.2}{sec:traces-scripts}

Each plan is scored with eighteen binary keyword features (Table~\ref{tab:trace-dictionary}), applied as transparent regular expressions. The features were tuned on the language models actually produce: an initial dictionary keyed to the theory's vocabulary (``no risk,'' ``can't lose'') turned out to be essentially absent from real plans---\texttt{safety\_\allowbreak recognition}, which detects an echo of the \textit{Payoff Safety} invariant, fires on $3$ of $21{,}990$ plans---so the safety and risk group was re-anchored on the words models use (``overpay,'' ``profit margin,'' ``buffer,'' ``zero profit''). The dictionary does no negation handling by design: \texttt{overpay\_\allowbreak concern} counts ``avoid overpaying'' as overpayment rhetoric either way, which is the intended reading for the main panel but a blind spot at the frontier (Section~C.8).

\begin{table}[htbp]
\revfloat
\centering
\footnotesize
\setlength{\tabcolsep}{5pt}
\begin{tabular}{llp{6.6cm}}
\toprule
Group & Feature & Fires when the plan\ldots \\
\midrule
Normative & \texttt{dominance\_\allowbreak language} & invokes dominance (``dominant strategy,'' ``regardless of others'') \\
recognition & \texttt{truthful\_\allowbreak intent} & states an intent to bid exactly one's value \\
& \texttt{payment\_\allowbreak rule\_\allowbreak correct} & correctly rehearses that the winner pays the second-highest bid \\
& \texttt{second\_\allowbreak price\_\allowbreak mention} & name-drops ``second-highest''/``second price'' (weak rule mention) \\
& \texttt{first\_\allowbreak price\_\allowbreak mention} & calls the auction ``first-price'' (mechanism confusion) \\
\addlinespace
Opponents \& & \texttt{opponent\_\allowbreak modeling} & speculates about other bidders' values or bids \\
beliefs & \texttt{probability\_\allowbreak reasoning} & uses chance/likelihood language \\
& \texttt{expected\_\allowbreak value\_\allowbreak reasoning} & computes or invokes expected value \\
\addlinespace
Stated intent & \texttt{shading\_\allowbreak intent} & states an intent to bid \emph{below} value (value-anchored) \\
& \texttt{overbid\_\allowbreak intent} & states an intent to bid \emph{above} value (value-anchored) \\
\addlinespace
Safety / risk & \texttt{worst\_\allowbreak case} & reasons about the worst case \\
rhetoric & \texttt{safety\_\allowbreak recognition} & echoes the \textit{Payoff Safety} invariant (``bid determines if you win, not what you pay'') \\
& \texttt{overpay\_\allowbreak concern} & worries about overpaying \\
& \texttt{zero\_\allowbreak profit\_\allowbreak fallacy} & argues that bidding one's value yields zero profit \\
& \texttt{margin\_\allowbreak language} & invokes a profit margin or buffer \\
\addlinespace
Style & \texttt{conservative\_\allowbreak language} & self-describes as cautious/conservative \\
& \texttt{aggressive\_\allowbreak language} & self-describes as aggressive/competitive \\
& \texttt{risk\_\allowbreak language} & uses risk vocabulary \\
\bottomrule
\end{tabular}
\caption{The frozen trace feature dictionary: eighteen binary keyword features in five groups. Exact regular expressions and per-cell prevalence are in the replication package (\texttt{analysis/\allowbreak build\_\allowbreak trace\_\allowbreak features.py}; \texttt{results/\allowbreak traces/\allowbreak trace\_\allowbreak features.csv}).}
\label{tab:trace-dictionary}
\end{table}

\paragraph{Decision scripts.}
A rule-based decision list (\texttt{analysis/\allowbreak classify\_\allowbreak heuristics.py}) over the features, the additive formal-derivation group, and a stated-bid parser assigns each plan one primary script: H1 profit-margin shading; H2 avoid overpayment; H3 opponent anchoring; H4 salient-number anchoring without a value-based rule; H5 win-probability or aggressive bidding; H6 price-threshold exit (clock); H7 formal dominance derivation; H8 worst-case safety proof; T informal truthful or dominance assertion; U unclassified. Because the LLM-judge replication (Section~C.3) shows that H1 and H2, and H7 and T, are not reliably separable from text alone, the main text reports shares at a coarse grain: below-value script (H1$+$H2), opponent (H3), anchor (H4), aggressive (H5), clock exit (H6), normative (H7$+$H8$+$T). Script labels are descriptive only and are never outcome variables in the intervention analysis of Section~C.5. Verbatim exemplars for every label are in \texttt{results/\allowbreak traces/\allowbreak heuristic\_\allowbreak examples.md}.

\paragraph{The six scripts, with exemplars.}
The plans are not cheap talk, but neither are they textbook reasoning about the second-price rule. Read by primary script (Figure~\ref{fig:heuristic-prevalence}), the modal plan in the sealed second-price auction is \emph{first-price} logic: a heuristic that would serve a bidder well in an auction where the winner pays her own bid, applied to a rule under which she does not. Six scripts account for nearly all classified plans; we give one verbatim exemplar of each.

\begin{itemize}
    \item \textit{Profit-margin shading}---bid below value to keep a margin between value and payment (GPT-4o, value \$34, bid \$33.9): ``Bid just below my value to maximize my profit if I win. Aim to outbid others while staying below \$34 to ensure positive earnings.'' The script uses own value and a desired margin and ignores the rule that sets the payment. Among plans stating an intent to shade, $97.1\%$ of realized bids fall below value (mean deviation $-\$1.84$). It is the primary script of $63\%$ of Gemini's sealed second-price plans, $38\%$ of GPT-4o's, and $37\%$ of Claude's, but only $2\%$ of Gemma's.
    \item \textit{Avoid overpayment}---lower the bid whenever a high bid feels capable of causing a loss (Gemini, value \$44, bid \$43.9): ``I will bid slightly below my value to ensure a positive profit if I win. I will bid \$43.9 to avoid overbidding and risking a loss.'' This is the failure to recognize the payoff-safety invariant in its purest form; it is Gemma's modal script ($38\%$) and GPT-4o's second ($28\%$).
    \item \textit{Opponent anchoring}---forecast competitors' bids and bid just above or below the forecast (GPT-4o, value \$33, bid \$31): ``To maximize profit, bid slightly above the expected average bid. Assume others bid around \$25--\$30. Bid \$31 to win while minimizing payment risk.'' The beliefs are unnecessary for the dominant-strategy comparison. This script is a minority one at baseline; it is the one the belief scaffolds of Appendix~\ref{sec:beliefs} activate.
    \item \textit{Salient-number anchoring without a value-based rule} (Gemini, value \$17, bid \$8.5): ``I will bid a fraction of my value. A good starting point is half my value. [\ldots] I will bid 8.5.'' Plans with no value-anchored intent at all carry the largest errors (Appendix~\ref{sec:traces-fidelity}).
    \item \textit{Win-probability, or aggressive bidding} (GPT-4o, value \$14, bid \$14.5): ``Bid slightly above my value to maximize profit if I win. Since my value is \$14, I will bid \$14.5 to increase chances of winning while minimizing potential loss.'' Among plans stating an intent to bid above value, $87.6\%$ do (mean $+\$1.88$); Claude is the outlier here, with an aggressive primary script in $30\%$ of its plans.
    \item \textit{Price-threshold exit}, the clock's script (Gemini, value \$34): ``My value is 34. I will stay in as long as the price is less than or equal to 33. I will drop out if the price reaches 33.5.'' We return to it in Appendix~\ref{sec:traces-clock}.
\end{itemize}

Two facts about these scripts matter for what follows. First, each family has a recognizable fingerprint---Gemini a near-pure shader, Gemma an unanchored worrier about overpaying, Claude the noisiest and most often aggressive, GPT-4o a shader with a secondary overpayment script---and the fingerprints line up with the ordering of baseline errors in Section~\ref{sec:osp-auctions}. Second, the scripts are largely \emph{insensitive to the payment rule}. In the cross-mechanism control cells, GPT-4o states an intent to shade in $66\%$, $68\%$, and $60\%$ of its first-, second-, and third-price plans and bids $0.82$--$0.88$ of value in all three; in a smaller per-format grid Gemini keeps its shading script across all three rules and Gemma's overpayment script is likewise mechanism-insensitive, while Claude is mixed (Appendix~\ref{sec:appendix-traces}). A script that is a reasonable heuristic in the first-price auction is carried, unchanged, into a rule where it is an error.

\begin{figure}[htbp]
    \revfloat
    \centering
    \includegraphics[width=0.95\textwidth]{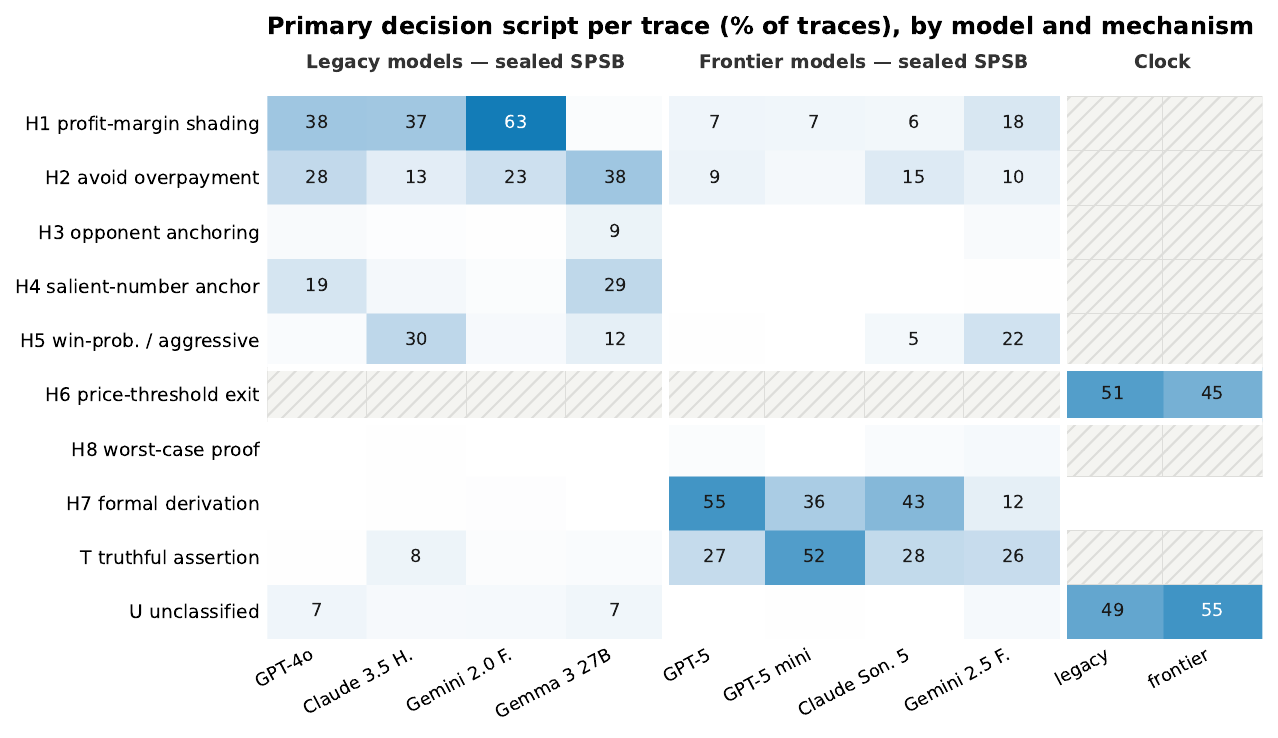}
    \caption{Primary decision script per plan, by model and mechanism (rule-based taxonomy; hatched cells are structurally unavailable labels). The plans of the four families in our main panel (``legacy'') state heterogeneous sealed-bid heuristics dominated by first-price shading and overpayment avoidance; the ascending clock collapses every family onto the price-threshold exit rule; the frontier families discussed in Appendix~\ref{sec:traces-implications} switch to formal derivation or truthful assertion. LLM-judge replication of the taxonomy: $\kappa = 0.55$ at the coarse grain shown; ten-label grain and confusion matrix in Appendix~\ref{sec:appendix-traces}.}
    \Description{Stacked bars of primary decision-script shares per model and mechanism.}
    \label{fig:heuristic-prevalence}
\end{figure}

\paragraph{Fingerprints.}
Table~\ref{tab:trace-prevalence-models} reports feature prevalence in the baseline SPSB cell of each model ($n = 150$ plans per model). The ``fingerprint'' shares quoted in Section~\ref{sec:traces-scripts} instead pool \emph{all} sealed second-price cells of a model (baseline and interventions), which is why adjacent numbers differ slightly (Gemini \texttt{shading\_\allowbreak intent} $88\%$ in the baseline cell versus $86\%$ pooled; Gemma $4\%$ versus $11\%$). Both granularities are internally consistent; the text labels which one it uses.

\begin{table}[htbp]
\revfloat
\centering
\footnotesize
\setlength{\tabcolsep}{6pt}
\begin{tabular}{lcccc}
\toprule
Feature (\% of plans) & Claude & Gemini & GPT-4o & Gemma \\
\midrule
\texttt{dominance\_\allowbreak language} & 0.0 & 0.7 & 0.0 & 0.0 \\
\texttt{truthful\_\allowbreak intent} & 8.0 & 2.7 & 2.0 & 0.0 \\
\texttt{payment\_\allowbreak rule\_\allowbreak correct} & 11.3 & 0.0 & 9.3 & 0.0 \\
\texttt{second\_\allowbreak price\_\allowbreak mention} & 50.7 & 10.0 & 39.3 & 46.0 \\
\texttt{first\_\allowbreak price\_\allowbreak mention} & 18.7 & 0.0 & 0.0 & 0.0 \\
\texttt{opponent\_\allowbreak modeling} & 8.0 & 4.7 & 20.0 & 73.3 \\
\texttt{probability\_\allowbreak reasoning} & 49.3 & 22.7 & 38.0 & 18.0 \\
\texttt{expected\_\allowbreak value\_\allowbreak reasoning} & 3.3 & 2.7 & 2.0 & 0.0 \\
\texttt{shading\_\allowbreak intent} & 54.0 & 88.0 & 61.3 & 4.0 \\
\texttt{overbid\_\allowbreak intent} & 25.3 & 0.0 & 2.7 & 2.0 \\
\texttt{worst\_\allowbreak case} & 6.0 & 8.0 & 2.7 & 1.3 \\
\texttt{safety\_\allowbreak recognition} & 0.0 & 0.0 & 0.0 & 0.0 \\
\texttt{overpay\_\allowbreak concern} & 59.3 & 22.7 & 45.3 & 89.3 \\
\texttt{zero\_\allowbreak profit\_\allowbreak fallacy} & 0.0 & 0.0 & 1.3 & 0.0 \\
\texttt{margin\_\allowbreak language} & 54.7 & 29.3 & 30.7 & 13.3 \\
\texttt{conservative\_\allowbreak language} & 6.0 & 14.7 & 12.7 & 86.0 \\
\texttt{aggressive\_\allowbreak language} & 2.7 & 5.3 & 2.0 & 17.3 \\
\texttt{risk\_\allowbreak language} & 52.0 & 6.7 & 32.7 & 22.7 \\
\midrule
Plans & 150 & 150 & 150 & 150 \\
\bottomrule
\end{tabular}
\caption{Feature prevalence in the baseline SPSB cell, by model (percent of plans; Claude 3.5 Haiku, Gemini 2.0 Flash, GPT-4o, Gemma 27B). Gemini is a near-pure shader; Claude is the only model that mislabels the mechanism ``first-price'' and has both the most stated overbidding and the most stated truthful intent; Gemma produces the least value-anchored language but the most opponent modeling, conservative self-description, and overpayment concern; GPT-4o shades and has the highest rate of the zero-profit fallacy. Prevalence for every model $\times$ lever cell: \texttt{results/\allowbreak traces/\allowbreak feature\_\allowbreak prevalence\_\allowbreak model\_\allowbreak family.csv}.}
\label{tab:trace-prevalence-models}
\end{table}

\paragraph{Mechanism-insensitive scripts.}
Table~\ref{tab:script-invariance} reports the shading and overpayment scripts by payment rule. GPT-4o, for which the full first-, second-, and third-price grids exist, states an intent to shade in $66\%$, $68\%$, and $60\%$ of plans and bids $0.82$--$0.88$ of value under all three rules---one script whether shading is optimal (first-price) or an error (second-price). In the smaller per-format robustness grid, Gemini keeps a below-value script across all three rules ($75$--$87\%$ H1 or H2 primary), Gemma's overpayment concern is likewise mechanism-insensitive ($71$--$88\%$), and Claude is mixed. A useful side effect is a validity check on the dictionary itself: \texttt{shading\_\allowbreak intent} prevalence is nearly identical where shading is correct and where it is an error, so the feature measures the stated strategy, not its appropriateness.

\begin{table}[htbp]
\revfloat
\centering
\footnotesize
\setlength{\tabcolsep}{5pt}
\resizebox{\linewidth}{!}{%
\begin{tabular}{llrccc}
\toprule
Model & Rule & Plans & \texttt{shading\_\allowbreak intent} & H1 or H2 primary & bid/value \\
\midrule
GPT-4o (main grid) & first-price & 3{,}711 & 66\% & 68\% & 0.82 \\
 & second-price & 3{,}600 & 68\% & 66\% & 0.88 \\
 & third-price & 3{,}696 & 60\% & 59\% & 0.88 \\
\addlinespace
Gemini 2.0 Flash (robustness grid) & first-price & 180 & 45\% & 75\% & 0.88 \\
 & second-price & 264 & 64\% & 80\% & 0.93 \\
 & third-price & 330 & 86\% & 87\% & 0.90 \\
\addlinespace
Gemma 27B (robustness grid) & first-price & 180 & 2\% & 17\% & 0.85 \\
 & second-price & 270 & 7\% & 48\% & 0.82 \\
 & third-price & 150 & 9\% & 55\% & 0.75 \\
\addlinespace
Claude 3.5 Haiku (robustness grid) & first-price & 177 & 40\% & 56\% & 0.98 \\
 & second-price & 267 & 18\% & 37\% & 1.07 \\
 & third-price & 330 & 46\% & 48\% & 1.07 \\
\bottomrule
\end{tabular}}
\caption{Cross-mechanism script invariance (\texttt{results/\allowbreak traces/\allowbreak script\_\allowbreak invariance.csv}). For Gemma, the mechanism-insensitive script is overpayment concern (\texttt{overpay\_\allowbreak concern} $88\%$, $84\%$, $71\%$ across the three rules), not shading. The rule-rehearsal features are mechanically zero outside the second-price format and are omitted.}
\label{tab:script-invariance}
\end{table}

\subsection*{C.3\quad LLM-Judge Validation}

A stratified sample of $2{,}573$ plans (stratified by corpus, model, and rule label; seed 1299) was independently labeled by an LLM judge (Claude Haiku 4.5) that saw only the plan text and the mechanism family---no bid, value, model, or rule label. Its role is a robustness check on taxonomy shares and on plans without a value-based bid rule; judge labels are never the outcome variable in the intervention analysis. Agreement with the rule-based label is Cohen's $\kappa = 0.36$ at the ten-label grain (raw agreement $0.43$), $\kappa = 0.45$ with H1$+$H2 and H7$+$T merged, and $\kappa = 0.55$ at the coarse decision-mode grain used in the text (raw agreement $0.64$; $0.55$ for the main panel and $0.52$ for the frontier corpus). The largest disagreements are informative: the judge folds most rule-labeled ``avoid overpayment'' plans into profit-margin shading or truthful assertion (H2 recall $0.06$), which is why the two below-value scripts are only ever reported merged; it reads nearly every rule-labeled ``unclassified'' clock rationale as the price-threshold exit rule (H6 recall $0.96$); and it relabels a large share of formal derivations as informal truthful assertions (H7 recall $0.32$), which is why the normative family is reported merged. The full confusion matrix is in \texttt{results/\allowbreak traces/\allowbreak judge/\allowbreak judge\_\allowbreak validation.md}.

The judge also carries a broader ``states why truthful bidding is safe'' flag, which cross-checks the frozen \texttt{safety\_\allowbreak recognition} regex (3 hits in $21{,}990$ plans). The flag fires on $2.7\%$ of main-panel plans (Claude $5.4\%$, Gemini $3.8\%$, GPT-4o $1.5\%$, Gemma $0.3\%$) and on $45\%$ of frontier plans (Claude Sonnet 5 $58\%$, GPT-5 $54\%$, GPT-5 mini $49\%$, Gemini 2.5 Flash $26\%$). The broader labels also indicate that explicit safety reasoning is uncommon in the main panel, although its measured prevalence depends on the definition. An optional $\sim$100-plan human hand-labeling, to complement the LLM judge, has a protocol in the same file and has not yet been run.

\subsection*{C.4\quad Stated Intent, Realized Bids, and the Pooled Regression}
\applabel{C.4}{sec:traces-fidelity}
If plans were noise, none of this would matter. They are not. Where a plan names a dollar amount, the realized bid matches it almost exactly: the correlation between stated and realized bid is $0.97$--$0.99$ across the four families, with a mean absolute gap of \$0.28--\$0.82, and stated direction (below, at, or above value) agrees with realized direction for $83$--$98\%$ of plans, three of four families above $97\%$ (Figure~\ref{fig:fidelity}a). The models do what they say. The error is in \emph{which plan gets selected}: conditional on the stated-intent class, the plans that never anchor on value at all---$4{,}541$ of them---carry by far the largest deviations (pooled mean $-\$4.84$; Figure~\ref{fig:fidelity}b), and they are not one phenomenon. Gemma contributes $2{,}613$ of them ($75\%$ of its own plans, mean deviation $-\$6.05$, three quarters of them self-described as ``cautious'' or ``conservative''), GPT-4o $1{,}040$ at $-\$4.41$, while Claude's $578$ unanchored plans sit at $+\$0.27$: the same label is Gemma's dominant failure mode and Claude's benign residual. A pooled regression of $|b-v|$ on the full dictionary, with model and lever fixed effects, adds an incremental $R^2$ of only $0.075$ ($0.069$ deduplicated): the plans are informative about \emph{how} an agent errs, far from a sufficient statistic for the error itself.

\begin{figure}[htbp]
    \revfloat
    \centering
    \includegraphics[width=\textwidth]{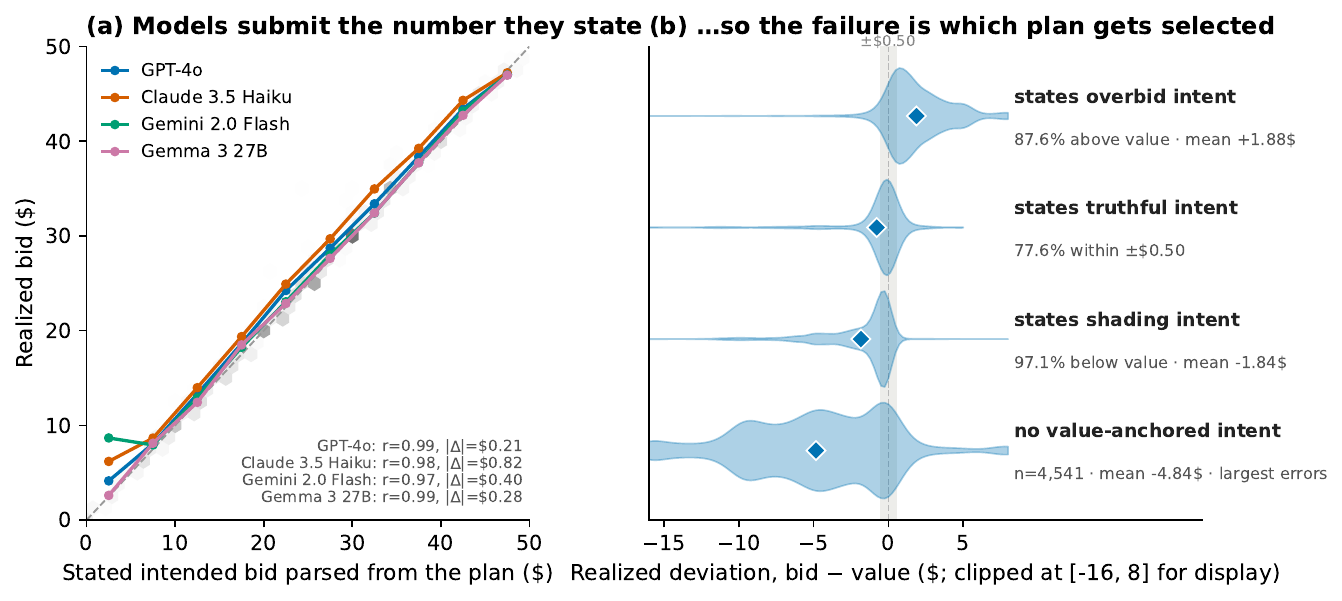}
    \caption{Execution fidelity versus plan selection (main panel, sealed second-price cells). (a)~Realized bid against the dollar amount stated in the plan; per-model binned means, $r = 0.97$--$0.99$, mean $|\text{stated} - \text{realized}|$ of \$0.28--\$0.82. (b)~Realized deviation conditional on the stated-intent class; the plans that never anchor on value carry the largest errors. Stated plans are executed faithfully; the failure is in which plan is chosen.}
    \Description{Scatter of realized versus stated bids and conditional deviation by stated-intent class.}
    \label{fig:fidelity}
\end{figure}

Among the $7{,}469$ sealed second-price plans stating an intent to shade, $97.1\%$ of realized bids fall below value (mean deviation $-\$1.84$); among the $1{,}797$ stating an intent to bid above value, $87.6\%$ fall above (mean $+\$1.88$); among the $950$ stating truthful intent, $77.6\%$ land within $\pm\$0.50$ of value (mean $|b-v|$ of \$1.03). The largest errors come from the $4{,}541$ plans that state no value-anchored intent at all: pooled mean deviation $-\$4.84$, mean absolute deviation \$5.51. The per-model autopsy (\texttt{results/\allowbreak traces/\allowbreak mediation/\allowbreak unanchored\_\allowbreak autopsy.csv}) shows this is not one phenomenon: Gemma contributes $2{,}613$ unanchored plans ($75\%$ of its own; mean deviation $-\$6.05$; $76\%$ self-described as conservative, $83\%$ voicing overpayment concern, $54\%$ speculating about opponents), GPT-4o $1{,}040$ ($29\%$; $-\$4.41$), Gemini $310$ ($9\%$; $-\$5.61$), and Claude $578$ ($17\%$; $+\$0.27$). Where a plan names a dollar amount---it does so in $80$--$93\%$ of main-panel plans---the realized bid matches it with correlation $0.97$--$0.99$ and a mean absolute gap of \$0.28 (Gemma, GPT-4o), \$0.40 (Gemini), and \$0.82 (Claude); stated direction matches realized direction in $97$--$98\%$ of plans for Claude, Gemini, and GPT-4o and $83\%$ for Gemma (\texttt{results/\allowbreak traces/\allowbreak mediation/\allowbreak consistency\_\allowbreak pooled\_\allowbreak v2.csv}).

A pooled OLS of $|b-v|$ on the full dictionary with model and lever-family fixed effects ($n = 14{,}073$; standard errors cluster-robust by run directory) summarizes the same facts: articulating \emph{any} value-anchored plan---shading, overbidding, or truthful---is associated with \$1.5--\$2.2 smaller absolute deviations than an unanchored plan, correct payment-rule rehearsal with a further $-\$0.77$, and conservative self-description is the one style feature with a robust positive (worse) association ($+\$0.76$). Overall explanatory power is modest: $R^{2} = 0.251$ with the features against $0.176$ with fixed effects only, an incremental $R^{2}$ of $0.075$ ($0.069$ on the deduplicated pool). Stated reasoning is informative about \emph{how} an agent errs, far from a sufficient statistic for the error itself.

\subsection*{C.5\quad Effects of All Twenty Interventions}

Table~\ref{tab:trace-interventions-full} reports all twenty interventions. For each, the targeted feature is the one the lever's text is designed to move; the language and bid columns give baseline and treated prevalence and mean $|b-v|$; the $q$-values are Benjamini--Hochberg corrected within two families (all twenty language tests; all twenty bid tests), computed from wild-cluster-bootstrap $p$-values clustered at the run level. The four primary contrasts are \emph{classified} at their raw $p$ (\textit{Payoff Safety} bids $p = 0.015$; \textit{Payoff Tree} bids $p = 0.017$; worst-case language $p = 0.0075$, bids $p = 0.81$; menu bids $p = 0.66$); the full-set $q$ is shown for completeness. Baselines are the pooled contingent-reasoning and belief baseline cells for levers whose prompts share the standard rule text ($|b-v| = 3.20$, $n = 1{,}188$); the contingent-reasoning grid's own baseline for its scaffolds ($3.24$, $n = 588$); the belief grid's own baseline for the belief prompts ($3.16$, $n = 600$); and the loss-aversion grid's baseline for its frames ($3.13$, $n = 567$). The legacy ``forward-planning baseline'' cell is excluded from every baseline pool and analyzed as a treatment (the two-stage clock-exit description), because its text is itself a clock-framed description of the auction.

\begin{table}[htbp]
\revfloat
\centering
\scriptsize
\setlength{\tabcolsep}{2.5pt}
\begin{tabular}{p{2.5cm}p{2.3cm}ccccllc}
\toprule
& & \multicolumn{2}{c}{Stated reasoning} & \multicolumn{2}{c}{Bids, mean $|b-v|$ (\$)} & & & \\
\cmidrule(lr){3-4}\cmidrule(lr){5-6}
Lever & Targeted feature & \% base $\to$ treated & $q$ & base $\to$ treated & $q$ & Echo & Sign & Cell \\
\midrule
Payoff Safety$^{\dagger}$ & \texttt{safety\_\allowbreak \allowbreak recognition} & $0.0 \to 0.0$ & --- & $3.20 \to 1.95$ & 0.068$^{*}$ & clean & \texttt{----} & bids-only \\
Worst-case scaffold$^{\dagger}$ & \texttt{worst\_\allowbreak case} & $2.0 \to 43.0$ & 0.020 & $3.24 \to 3.07$ & 0.93 & partial & \texttt{-+-+} & language-only \\
Payoff Tree$^{\dagger}$ & \texttt{second\_\allowbreak \allowbreak price\_\allowbreak \allowbreak mention} & $33.0 \to 32.7$ & 0.91 & $3.20 \to 1.29$ & 0.068$^{*}$ & clean & \texttt{----} & bids-only \\
Dominated & \texttt{dominance\_\allowbreak \allowbreak language} & $0.0 \to 20.5$ & 0.035 & $3.24 \to 2.88$ & 0.93 & partial & \texttt{-+-+} & language-only \\
Enumerate & \texttt{opponent\_\allowbreak \allowbreak modeling} & $29.1 \to 39.3$ & 0.84 & $3.24 \to 3.28$ & 0.93 & partial & \texttt{-+-+} & neither \\
Menu restatement$^{\dagger}$ & \texttt{second\_\allowbreak \allowbreak price\_\allowbreak \allowbreak mention} & $33.0 \to 0.0$ & 0.015 & $3.20 \to 3.55$ & 0.93 & by constr. & \texttt{+--+} & neither \\
Clock-framing & \texttt{clock\_\allowbreak \allowbreak exit\_\allowbreak \allowbreak language} & $0.0 \to 19.5$ & 0.023 & $3.20 \to 2.24$ & 0.59 & manip. & \texttt{-+--} & language-only \\
Two-stage clock-exit description & \texttt{clock\_\allowbreak \allowbreak exit\_\allowbreak \allowbreak language} & $0.0 \to 14.3$ & 0.015 & $3.20 \to 3.41$ & 0.93 & manip. & \texttt{++--} & language-only \\
First-order beliefs & \texttt{opponent\_\allowbreak \allowbreak modeling} & $27.8 \to 42.7$ & 0.015 & $3.16 \to 3.93$ & 0.027$^{*}$ & partial & \texttt{++++} & both \\
Second-order beliefs & \texttt{opponent\_\allowbreak \allowbreak modeling} & $27.8 \to 51.3$ & 0.020 & $3.16 \to 3.87$ & 0.36 & partial & \texttt{+-++} & language-only \\
Common-knowledge beliefs & \texttt{opponent\_\allowbreak \allowbreak modeling} & $27.8 \to 11.8$ & 0.27 & $3.16 \to 2.13$ & 0.027$^{*}$ & partial & \texttt{----} & bids-only \\
Backward induction & \texttt{expected\_\allowbreak \allowbreak value\_\allowbreak \allowbreak reasoning} & $4.4 \to 4.0$ & 0.84 & $3.20 \to 2.23$ & 0.36 & partial & \texttt{+---} & neither \\
Risk-averse persona & \texttt{risk\_\allowbreak \allowbreak language} & $29.0 \to 99.3$ & 0.015 & $3.20 \to 4.71$ & 0.027$^{*}$ & partial & \texttt{++++} & both \\
Risk-neutral persona & \texttt{risk\_\allowbreak \allowbreak language} & $29.0 \to 46.8$ & 0.30 & $3.20 \to 2.57$ & 0.23 & partial & \texttt{----} & neither \\
Risk-seeking persona & \texttt{aggressive\_\allowbreak \allowbreak language} & $5.8 \to 65.5$ & 0.023 & $3.20 \to 3.49$ & 0.93 & partial & \texttt{+---} & language-only \\
Loss frame & \texttt{risk\_\allowbreak \allowbreak language} & $32.1 \to 35.8$ & 0.84 & $3.13 \to 3.73$ & 0.93 & partial & \texttt{-+++} & neither \\
Gain frame & \texttt{risk\_\allowbreak \allowbreak language} & $32.1 \to 26.5$ & 0.84 & $3.13 \to 3.06$ & 0.93 & partial & \texttt{+--+} & neither \\
Endowment & \texttt{risk\_\allowbreak \allowbreak language} & $32.1 \to 35.2$ & 0.84 & $3.13 \to 4.00$ & 0.10 & partial & \texttt{-+++} & neither \\
Mixed frame & \texttt{risk\_\allowbreak \allowbreak language} & $32.1 \to 45.2$ & 0.015 & $3.13 \to 3.68$ & 0.93 & partial & \texttt{-+-+} & language-only \\
WTA/WTP & \texttt{risk\_\allowbreak \allowbreak language} & $32.1 \to 30.8$ & 0.91 & $3.13 \to 2.46$ & 0.59 & partial & \texttt{-+--} & neither \\
\bottomrule
\end{tabular}
\caption{The full set of interventions, pooled over the four families (sealed second-price cells). Primary contrasts (\textit{Payoff Safety}, \textit{Payoff Tree}, worst-case scaffold, menu restatement) are classified at their raw wild-cluster-bootstrap $p$-values (see text); all other rows at the full-set $q$ shown. $^{*}$: significant by the run-level wild-cluster bootstrap but soft under the four-model-cluster permutation test. \emph{Echo}: ``clean'' means the targeted vocabulary does not appear in the prompt; ``partial'' means part of the language response is the prompt's own vocabulary read back; ``by constr.'' means the menu wording removes the targeted vocabulary by construction, so its language column is not a comprehension measure; ``manip.'' means the language \emph{is} the manipulation (the two clock-framed descriptions), so it is not a comprehension measure either. $^{\dagger}$: primary contrast. \emph{Sign}: the direction of the $|b-v|$ change in Claude, Gemini, Gemma, GPT-4o ($-$ improves, $+$ worsens; \texttt{dissociation\_\allowbreak per\_\allowbreak family.csv}, auction-level clusters, descriptive since there is one run per model $\times$ cell). \emph{Cell}: joint-response class. Every quantity is stable on the deduplicated pool.}
\label{tab:trace-interventions-full}
\end{table}

Three rows beyond those in the main text deserve a note. The \emph{Dominated} scaffold (``identify and eliminate dominated bids'') is a third words-only lever: dominance language rises from $0\%$ to $20.5\%$ ($q = 0.04$) while bids do not move. The \emph{clock-framing} description and the \emph{two-stage clock-exit} description both pass their manipulation checks (clock-and-exit language $0 \to 20\%$ and $0 \to 14\%$), with pooled bid effects that are null and conceal offsetting family effects (Gemma strongly helped, Gemini strongly harmed in both). And the \emph{common-knowledge} belief prompt is the one belief scaffold that improves bids ($3.16 \to 2.13$, $q = 0.027$, soft), while \emph{reducing} opponent-modeling language---consistent with its text, which asserts a conclusion (``all rational bidders would converge'') rather than inviting a forecast.

\subsection*{C.6\quad Mediation Bounds}

For each primary contrast we ask how much of the bid effect could run through verbalized understanding, using the classical product-of-coefficients decomposition with the normative-recognition features as mediators (\texttt{results/\allowbreak traces/\allowbreak mediation/\allowbreak mediation\_\allowbreak bounds\_\allowbreak all.csv}). For \textit{Payoff Safety} the first stage is null on every measure: the invariant the description asserts is echoed in $0$ of $600$ treated plans (rule-of-three one-sided 95\% upper bound on prevalence $0.5\%$), correct payment-rule rehearsal moves from $3.9\%$ to $2.8\%$ (TOST-equivalent to no change within $\pm 5$ points at both run- and auction-level clustering), and rule name-dropping from $33.0\%$ to $28.2\%$ ($p = 0.61$). The average causal mediation effect has a median of $0.02$ (95\% CI $[-0.17, 0.15]$) against a total effect of $-1.25$ on $|b-v|$, so the mediated share has a point estimate of roughly zero and a 95\% upper bound of $35.4\%$ in absolute value under the decomposition. This is conditional on its mediation assumptions and the chosen verbal indicators, rather than a bound on unobserved understanding. For the \textit{Payoff Tree} the first stage is $-0.4$ percentage points and the bound is $14.7\%$. For the worst-case scaffold the first stage is large ($+40.9$ points of worst-case language) but the total effect is null ($-0.17$), and within the treated group plans that articulate the worst case err no less than plans that do not (gap $+\$0.07$). Two cross-sectional associations are reported for completeness and are not evidence of mediation: within the \textit{Payoff Safety} cell the minority of plans that do rehearse the payment rule err $\$1.40$ less (cluster-bootstrap 95\% CI $[-2.15, -0.56]$), and in the trace-aligned logs Claude's \textit{Payoff Safety} cell has a mean signed deviation of $+0.30$ (mild overbidding), the value used in Section~\ref{sec:mech-description}.

\subsection*{C.7\quad Robustness and Measurement Caveats}

\paragraph{Byte-identical duplicates.} Roughly $40\%$ of sealed-bid plans are byte-identical duplicates within model $\times$ cell $\times$ value draw: at temperature $0.5$ the models are near-deterministic conditional on the value, and every duplicate group shares a single value and a single bid. Full-sample standard errors are therefore optimistic, and the effective sample per cell is closer to the number of distinct value draws than to the row count. Re-running the pooled regression on deduplicated data ($n = 8{,}839$) leaves every headline coefficient stable (\texttt{shading\_\allowbreak intent} $-2.06 \to -1.90$, \texttt{overbid\_\allowbreak intent} $-2.18 \to -2.17$, \texttt{truthful\_\allowbreak intent} $-1.50 \to -1.46$, \texttt{payment\_\allowbreak rule\_\allowbreak correct} $-0.77 \to -0.77$, \texttt{conservative\_\allowbreak language} $+0.76 \to +0.72$; all $p \le 0.001$), and every intervention keeps its class.

\paragraph{Cluster counts.} The harmonized grid provides 92 run-directory clusters, but the menu and clock-framing cells are stored as one flat directory per model $\times$ cell, so pooled contrasts involving them rest on 8--12 clusters and their run-level $p$-values should be read accordingly; the four-model-cluster permutation test is the conservative cross-check, and effects it does not support are flagged soft. Per-family intervals use auction-level clusters and are descriptive, since there is one run per model $\times$ cell.

\paragraph{Negation.} The frozen intent regexes have no negation handling. This is harmless for the main panel, whose plans almost never discuss shading in order to reject it, but frontier derivations do (``shading only increases the chance of losing''), which fires \texttt{shading\_\allowbreak intent}: $198$ of GPT-5's $558$ intent-stated plans are scored as direction mismatches this way, yet $100\%$ of them bid within $\pm\$0.50$ of value. Frontier direction-consistency figures are therefore artifacts of this blind spot; the dollar-level fidelity figures (GPT-5 and GPT-5 mini state and submit the same dollar amount with exact match $1.00$) are not.

\paragraph{Clock logs.} Clock logs store the bidding history in exit order, not agent order; the corpus build realigns each rationale by agent name and flags the winner, whose recorded price is censored at the runner-up's exit. An earlier positional parse exaggerated frontier clock error by an order of magnitude; any clock number computed without the alignment is invalid. With the alignment, $34$--$85\%$ of main-panel clock plans state the threshold rule verbatim enough for the strict regex (Claude $47\%$, Gemini $34\%$, Gemma $41\%$, GPT-4o $85\%$), losing bidders' exits sit within \$0.33--\$1.16 of value, and no bidder in any main-panel or frontier family stays past its value (\texttt{results/\allowbreak traces/\allowbreak clock\_\allowbreak h6.csv}).

\subsection*{C.8\quad The Frontier Boundary}
\applabel{C.8}{sec:traces-implications}
The plans also mark a boundary for our main panel. We excluded frontier reasoning models from the panel on the grounds that raw reasoning ability may mask the bottlenecks our interventions target (Section~\ref{sec:methods}); their plans show what that masking looks like. Scoring $5{,}985$ plans from GPT-5, GPT-5 mini, Claude Sonnet 5, and Gemini 2.5 Flash with the same dictionary plus one additive feature group for formal derivation, a formal dominance argument---``bidding your true value is weakly dominant; overbidding risks winning when the second-highest bid exceeds your value''---is the primary script in $55\%$, $36\%$, $43\%$, and $12\%$ of their sealed plans respectively, against at most $1.2\%$ in any family of our panel, and the judge's ``states why truthful bidding is safe'' flag fires on $45\%$ of frontier plans against $2.7\%$ of ours. In these plans the decision is written up as a solved mathematics problem, a pattern that invites comparison with reasoning-oriented post-training studied in other models \citep{openai2024o1, guo2025deepseek}. This comparison does not identify the training history or cause of the pattern in the models tested here. That mode is a distinct channel, not a guarantee of reliability: conditional on a formal derivation GPT-5 and GPT-5 mini have zero bid error, but Claude Sonnet 5's error conditional on the same mode is \$1.55, driven entirely by $145$ plans that formally derive the \emph{wrong mechanism's} equilibrium---``the symmetric Nash equilibrium bidding strategy in a first-price auction is to bid $(n-1)/n$ of my value''---concentrated in the menu-restatement cell, which collapses Sonnet 5 from exact truthfulness to a mean $|b-v|$ of $8.7$ (and Gemini 2.5 Flash to $2.2$) while GPT-5 and GPT-5 mini are immune. A formally correct derivation of the wrong mechanism is the frontier's version of the script problem. The dictionary's negation blind spot (Section~C.7) qualifies frontier direction-consistency figures but not the dollar-level results below.

Table~\ref{tab:frontier-scripts} quantifies the shift in decision script between the main panel and four frontier models, scored with the frozen dictionary plus one additive formal-derivation feature group (equilibrium formula, dominance proof, environment recital). Formal dominance derivation is the primary script in $55\%$ of GPT-5's, $43\%$ of Claude Sonnet 5's, $36\%$ of GPT-5 mini's, and $12\%$ of Gemini 2.5 Flash's sealed second-price plans, with a further $26$--$52\%$ informal truthful assertion, against at most $1.2\%$ formal derivation in any main-panel family. Conditional on a formal derivation, GPT-5 and GPT-5 mini have zero bid error; Claude Sonnet 5's conditional error is \$1.55, driven entirely by $145$ plans that formally derive the \emph{first-price} equilibrium (``bid $(n-1)/n$ of my value''), with a mean deviation of $-\$8.81$. Those plans are concentrated in the menu-restatement cell, which collapses Sonnet 5 from exact truthfulness to mean $|b-v|$ of $8.7$ ($p = 0.0005$, auction-level clusters) and Gemini 2.5 Flash from $0.12$ to $2.24$, while GPT-5 and GPT-5 mini are unaffected. Description wording matters at the frontier in the clock as well: Claude Sonnet 5's exits are four times farther from value under the affiliated-values wording of the clock prompt than under the clean IPV wording (mean $|b-v|$ $2.12$ versus $0.49$; \texttt{results/\allowbreak traces/\allowbreak clock\_\allowbreak h6\_\allowbreak by\_\allowbreak experiment.csv}). The other primary interventions have nothing to improve at the frontier: baseline $|b-v|$ is $0.03$ pooled, and \textit{Payoff Safety}, \textit{Payoff Tree}, and the clock-framed descriptions leave it there.

\begin{table}[htbp]
\revfloat
\centering
\scriptsize
\setlength{\tabcolsep}{4pt}
\begin{tabular}{llrccccc}
\toprule
Model & Corpus & Plans & Formal (\%) & Truthful (\%) & $|b-v|$, formal & $|b-v|$, other & Wrong mech. \\
\midrule
Claude Sonnet 5 & frontier & 1,650 & 42.6 & 28.0 & 1.55 & 0.00 & 145 \\
Gemini 2.5 Flash & frontier & 1,608 & 11.8 & 25.8 & 0.20 & 0.61 & 9 \\
GPT-5 & frontier & 603 & 54.6 & 26.9 & 0.00 & 0.00 & 0 \\
GPT-5 mini & frontier & 1,374 & 35.7 & 51.9 & 0.00 & 0.00 & 0 \\
Claude 3.5 Haiku & main panel & 3,513 & 0.7 & 8.3 & 1.59 & 2.27 & 11 \\
Gemini 2.0 Flash & main panel & 3,486 & 1.2 & 1.6 & 3.68 & 1.51 & 1 \\
Gemma 3 27B & main panel & 3,474 & 0.0 & 2.8 & --- & 5.53 & 0 \\
GPT-4o & main panel & 3,600 & 0.0 & 0.7 & --- & 3.01 & 0 \\
\bottomrule
\end{tabular}
\caption{Decision scripts at the frontier versus the main panel, sealed second-price plans (\texttt{results/\allowbreak traces/\allowbreak formal\_\allowbreak derivation\_\allowbreak prevalence.csv}). ``Formal'' is the share of plans whose primary script is a formal dominance derivation (H7); ``Truthful'' the share with an informal truthful/dominance statement (T); ``$|b-v|$, formal/other'' the mean absolute deviation conditional on a formal derivation or on its absence. ``Wrong mech.'' counts plans that formally derive another mechanism's equilibrium (for Claude Sonnet 5, the first-price Nash bid). Conditional $|b-v|$ is in dollars; ``---'' means no plan of that kind.}
\label{tab:frontier-scripts}
\end{table}

Two further frontier facts bear on the main text. First, at the frontier the judge's ``states why truthful bidding is safe'' flag fires on $45\%$ of plans (Section~C.3), so the dissociation of Section~\ref{sec:traces-probe} is a statement about the main panel: models that already articulate the safety argument have nothing for \textit{Payoff Safety} to add, in words or in bids. Second, two exploratory frontier cells that test a \emph{false} safety statement (\texttt{pilot\_\allowbreak false\_\allowbreak safety\_\allowbreak up}/\texttt{\_\allowbreak down}) are computed but remain gated pending co-author review and are not part of any claim in this paper.
\subsection*{C.9\quad The Clock: The Environment Does the Work}
\applabel{C.9}{sec:traces-clock}
The ascending clock offers a clean check on the same dissociation from the other side. Between $34\%$ and $85\%$ of the clock plans state the exit-threshold rule verbatim enough for a strict pattern (the LLM judge reads essentially all clock rationales as this rule), exit prices sit within \$0.33--\$1.16 of value for losing bidders in every family, and no bidder in any family stays materially past its value. But exit fidelity is equally good whether or not the rule is stated: for Claude, mean $|b-v|$ is $0.27$ among plans that articulate the threshold and $0.43$ among those that do not.\footnote{Clock logs record the bidding history in exit order rather than agent order; the corpus realigns each rationale by agent name and treats the winner's price as censored at the runner-up's exit. An earlier positional parse overstated clock error by an order of magnitude.} The extensive form, not the verbalization, carries the improvement of Section~\ref{sec:osp-auctions}: the clock collapses every family onto one script (Figure~\ref{fig:heuristic-prevalence}) and makes that script safe to follow whether or not the agent can say why.
\end{revblock}

\section{Additional Experimental Details} \label{sec:appendix-details}

\rev{This appendix collects the environment details of Section~\ref{sec:methods} that are not needed to follow the contingent-reasoning and mechanism-description results; prompt skeletons are in Appendix~\ref{sec:appendix-prompts}.}

\subsection*{D.1\quad Deferred Acceptance: Market Instances, Priorities, and Protocols}
\paragraph{Market instances.}
Each instance contains students $i\in\{A,B,C,D\}$ and schools $s\in\{w,x,y,z\}$, with unit capacity (one seat per school). This $4\times 4$ set-up is large enough to admit meaningful strategic considerations (competition for popular schools and priority asymmetries) while remaining small enough that we can log the full DA trace and audit every outcome.

\paragraph{Student valuations.}
To induce correlated demand across students while preserving idiosyncratic tastes, we use an affiliated value model with a school-level common component and a student--school private shock:
\[
v_{i,s} \;=\; c_s \;+\; \varepsilon_{i,s},
\]
for school $s$ and student $i$.
For each school $s$, we draw a common value $c_s \sim \mathrm{Unif}[40,70]$. For each pair $(i,s)$, we draw an independent private shock $\varepsilon_{i,s}\sim \mathrm{Unif}[0,20]$ and set $v_{i,s}=c_s+\varepsilon_{i,s}$. Each student's \emph{true preference ranking} $\pi_i^{\star}$ is then defined as the descending sort of $\{v_{i,s}\}_{s\in\{w,x,y,z\}}$ (ties broken deterministically) \citep{klijn2019static}.

\paragraph{School priorities.}
Schools rank students by fixed, Ergin-acyclic priority orders (held constant across repetitions). Concretely, we use:
\begin{align*}
w:&\; A \succ B \succ C \succ D \\
x:&\; B \succ A \succ C \succ D \\
y:&\; A \succ B \succ D \succ C \\
z:&\; B \succ A \succ D \succ C,
\end{align*}
where $\succ$ denotes higher priority. This structure creates a ``top tier'' $\{A,B\}$ and ``bottom tier'' $\{C,D\}$ and satisfies the unit-capacity acyclicity condition. 
This acyclicity is the condition under which a sequential-query implementation can make DA's incentives more transparent (OSP).

\paragraph{Social information (common signal).}
In addition to their school priorities, all students observe a fixed ``global popularity'' ranking intended to proxy common beliefs about demand:
\[
\text{Global ranking: } y \succ x \succ w \succ z.
\]
This signal is held fixed across repetitions to keep the information treatment constant; depending on the realized $v_{i,s}$, it may be aligned or misaligned with the instance's true preference. 

\paragraph{Mechanisms.}
For each market instance we run two elicitation protocols.

\emph{(i) Direct revelation (static DA).} Each student submits a complete rank-order list over $\{w,x,y,z\}$ in a single shot. We then run student-proposing DA on the submitted rankings and with the true school priorities.

\emph{(ii) Iterative DA (sequential queries).} Students do not submit a full ranking upfront. Instead, the mechanism queries students sequentially, following the tree construction of \citet{ashlagi2018stable} for OSP implementation of DA under acyclic priorities. At each node, the mechanism identifies students who hold top priority at some remaining school and asks them a yes/no question: ``Among the remaining schools $\{w, x, y, z\}$, is school $s$ your most preferred?'' If the student answers YES, they are immediately matched to that school and both are removed from consideration. If NO, the school is not immediately removed and the mechanism continues to the next query; it may be assigned to another student before that student acts again. When only one top-priority student remains, the protocol reduces to serial dictatorship: the student simply picks their most preferred school from the remaining set. The protocol terminates when all students are matched.

\paragraph{Repetitions and logging.}
For each treatment condition and each mechanism, we target $N=50$ independent repetitions (distinct random seeds for value draws; priorities and global ranking fixed). For every repetition we log: true rankings, the full set of LLM responses (raw text), parsed decisions, DA traces (proposals/holds/rejections by round), final matches. Successful instances produce JSON records of value generation and matching outcomes. Retained cell sizes can be smaller after response or parsing failures; the sequential-query sample in Table~\ref{tab:da-decision-errors} uses the retained logs.

\subsection*{D.2\quad Auctions: Environment and Mechanisms}
\paragraph{Environment: Independent Private Values (IPV).}
We study symmetric IPV auctions with $N=3$ bidders. In each auction instance, each bidder $i\in\{1,2,3\}$ receives an independent private value
\[
v_i \sim \mathrm{Unif}\{0,1,\dots,V_{\max}\},
\]
with $V_{\max}=49$ in the baseline configuration. Values are drawn independently across bidders. All monetary amounts lie on a discrete grid with increment $\Delta=\$0.01$ for Sealed-bid and $\Delta=\$0.5$ for clock auctions.

\paragraph{Mechanisms.}
We implement two canonical one-shot mechanisms that differ in their strategic interface.

\emph{(i) Second-Price Sealed-Bid (SPSB).}
Each bidder simultaneously submits a bid $b_i \in \{0,\Delta,2\Delta,\dots\}$. The highest bidder wins the item and pays the second-highest bid. Let $b_{(1)} \ge b_{(2)} \ge b_{(3)}$ denote the order statistics of bids. The winner is the argmax bidder and the price is $p=b_{(2)}$. The winner’s payoff is $v_i-p$, and losers receive $0$.

\emph{(ii) Ascending Clock Auction (closed).}
The auction starts at price $p_0=0$ and increases by $\Delta$ each tick. At each posted price, each active bidder is asked whether they want to \emph{stay} in the auction or \emph{exit}. Exit decisions are private: bidders are not informed when others leave. The auction ends when only one bidder remains. We record each bidder’s \emph{dropout price} $d_i$ as the first posted price at which they choose to exit. The winner is the bidder with the highest dropout price, and the transaction price is the \emph{second-highest} dropout price (equivalently, the price at which the second-to-last bidder exits). The winner’s payoff is $v_i-\max_{j\neq i} d_j$, and losers receive $0$.

\section{Additional Results} \label{sec:appendix-results}

\rev{This appendix supplies the comparisons summarized in the main text: human benchmarks for the interface changes, followed by prompts that ask agents to plan through matching rounds or reason about other participants. The tasks and scoring remain those of Section~\ref{sec:methods}; lower absolute bid deviations and lower ranking errors indicate better play.}

\subsection*{E.1\quad Interface Comparisons: Human Benchmarks and Matching Error Measures}
\applabel{E.1}{sec:appendix-osp-more}
It is well known that truthful bidding is the dominant strategy in the SPSB auction \citep{vickrey1961counterspeculation}, yet human subjects systematically deviate from this in laboratory experiments \citep{kagel1993independent, kagel1995individual}. With regards to OSP mechanisms, \citet{li2017obviously} provides experimental evidence that an obviously strategy-proof clock format improves human bidding. \citet{breitmoser2022obviousness} decomposes features of ascending auctions: seeing the ascending clock and receiving dropout information improve bidding, while dynamic bidding itself has no significant incremental effect. These findings motivate separating interaction format, information, and description rather than attributing every clock-format gain to OSP alone.

In particular, if agents fail to appreciate that DA is strategy-proof, they may mistakenly misreport their rank-order list of preferences in seeking to improve their outcome given the reports of others. This misreport, then, is a function also of their beliefs of other agents' choice of what reports to make, and thus also their beliefs. To probe this, and also for realism, our implementation also reports to applicants that ``most other applicants seem to favor ${{global\_ranking}}$.'' Such beliefs, of course, should not change an applicant's report in DA if they understand the mechanism is strategy-proof. We test this question in Appendix~\ref{sec:beliefs}; full prompts are in Appendix~\ref{sec:appendix-prompts}.

\paragraph{Ranking summaries and decision errors.}
\rev{A complete submitted ranking reveals all six school pairs in a four-school market. Sequential responses reveal less: a single selected school has no rankable pair, and a false yes/no answer may not appear in the reconstructed ranking at all. The plotting pipeline assigns zero Kendall distance to a sequence with fewer than two ranked schools. Thus the zero recorded iterative statistic in Figure~\ref{fig:osp-comparison} is not evidence that every response was truthful.}

\rev{A separate analysis of the recorded sequential-query histories (the \texttt{osp\_\allowbreak yesno\_\allowbreak fixed} condition) scores each binary answer and each nonforced final choice against the values available at that node. NO is an error if the offered school is the unique highest-valued option; YES is an error if a strictly better school remains. A final choice is an error if its value is below the maximum available value. Choices between equal-valued best schools are accepted, and final choices with only one available school are excluded from the informative denominator. The resulting decision-level rates are shown in Table~\ref{tab:da-decision-errors}; they are not directly comparable to full-ranking Kendall distances.}

\begin{table}[htbp]
\centering
\small
\begin{tabular}{lrrrr}
\toprule
Model & Informative decisions & Errors & Error rate & False NO / YES \\
\midrule
Claude & 194 & 16 & $8.2\%$ & 13 / 2 \\
Gemini & 179 & 0 & $0.0\%$ & 0 / 0 \\
GPT-4o & 188 & 0 & $0.0\%$ & 0 / 0 \\
Gemma & 210 & 3 & $1.4\%$ & 3 / 0 \\
\bottomrule
\end{tabular}
\caption{Decision errors in the sequential-query DA logs. Informative decisions include yes/no answers and final choices among at least two schools. The last column counts false rejections and false acceptances; Claude's remaining error is a final-choice error. These are descriptive rates: decisions from the same market are dependent, and zero observed errors does not establish a zero population error rate.}
\label{tab:da-decision-errors}
\end{table}

\rev{The replication files also contain an earlier round-based choice protocol, \texttt{osp\_\allowbreak baseline}, which should not be conflated with the binary-query protocol. It has zero recorded Kendall distance for all four models, but decision-level scoring finds five strict-value choice errors among Gemma's 322 informative decisions and zero among the other three models' 346, 351, and 325 decisions. This further illustrates why zero partial-ranking distance does not imply zero decision errors. These existing-log checks change the interpretation of the sequential comparison, not the complete-ranking treatment comparisons ($4.2\%$ baseline, $1.7\%$ payoff tree, $0.2\%$ rejection safety).}

\subsection*{E.2\quad Forward Planning}
\applabel{E.2}{sec:forward}
The next axis we consider is that of forward planning. \citet{pycia2023theory} formalize a notion of limited forward planning that has become central to the theory of simple mechanisms. In a \emph{one-step simple} mechanism, an agent can identify the prescribed action by considering at most one of her own future moves at a time. In the ascending clock, for example, staying now can be justified by a plan to exit at the next opportunity; that plan can be updated when the next decision arrives. The agent need not plan the entire future sequence of prices. 

Both SPSB and DA are static mechanisms, with only a single decision node for each agent. But DA's underlying algorithm processes the submitted ranking dynamically, through multiple rounds of proposals and rejections, giving forward planning a natural place to bind: models may attempt to reason through the algorithm's rounds when deciding what ranking to submit. The sealed-bid auction has no analogous internal dynamics---a bid is submitted and the outcome is determined. In this sense, we think of DA as a dynamic \textit{algorithm}, and so test the forward planning interventions only on it. We test two `lookahead' scaffolds that prompt models to simulate the algorithm forward.

\begin{itemize}
    \item \emph{One-Step Lookahead}: ``If you are rejected from your first choice, which school would you propose to next? Does this affect how you should rank schools?''
    \item \emph{Two-Step Lookahead}: ``Consider the first two potential rejections. If rejected from your first choice, you go to your second. If rejected again, you go to your third. Plan your ranking with these steps in mind.''
\end{itemize}

\begin{figure}[htbp]
    \centering
    \includegraphics[width=\textwidth]{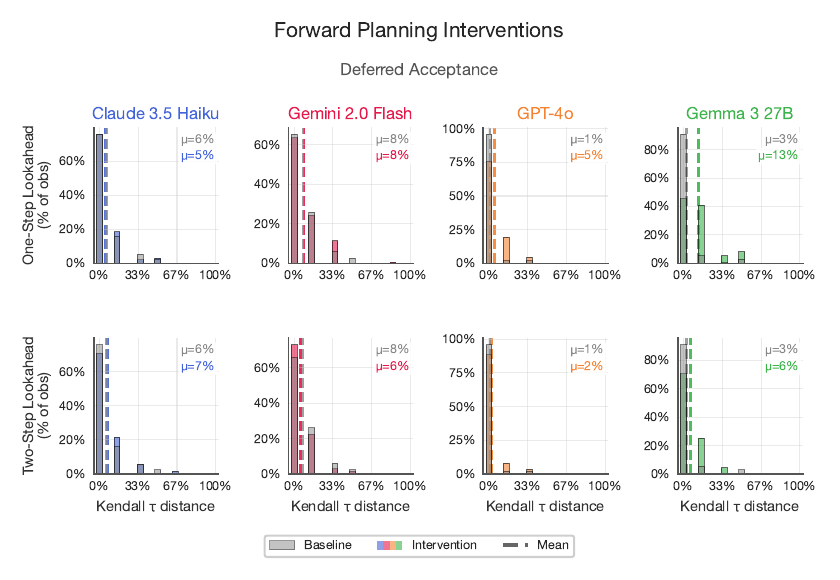}
    \caption{Forward planning: lookahead interventions in deferred acceptance. {\em Top row:} One-step lookahead; {\em bottom row:} two-step lookahead. Each cell overlays the direct DA baseline (gray) with the intervention (colored). Lookahead scaffolds worsen play overall, with the damage concentrated in GPT-4o and Gemma; Claude and Gemini are largely unaffected.}
    \label{fig:forward}
\end{figure}

While contingent reasoning was quite successful across models and mechanisms, it appears that forward planning scaffolds are less effective in improving play. The \textit{One-Step} intervention worsens play in DA from $\mu = 4.2\%$ to $7.8\%$ across models; \textit{Two-Step} worsens it to $5.1\%$. The damage is concentrated in the last two rows of GPT-4o and Gemma: \textit{One-Step} moves GPT-4o from $1.0\%$ to $4.8\%$ and Gemma from $2.6\%$ to $12.7\%$ error, whereas Claude and Gemini are largely unaffected. 

Why does prompting forward planning hurt? Our hypothesis is that it activates strategic thinking about the algorithm without resolving it. If models recognized DA is strategy-proof, forward planning would be irrelevant---the dominant strategy does not depend on what happens after a rejection. But because models do not fully appreciate strategy-proofness, prompting them to simulate rejection dynamics gives them material to strategize with, second-guessing their truthful preferences. The weaker models, with less ability to resolve the reasoning they have been prompted to undertake, suffer the most.

\subsection*{E.3\quad Belief Formation}
\applabel{E.3}{sec:beliefs}
The third axis is belief formation. One canonical story here is from \citet{borgers2019strategically}, who introduce strategic simplicity: a mechanism is said to be \textit{strategically simple} if optimal play does not depend on higher-order beliefs about what opponents believe. Both SPSB and DA are strategy-proof, and hence trivially strategically simple: the dominant strategy is independent of beliefs entirely. Scaffolding belief formation in these settings should therefore be irrelevant at best. But if models are already treating these mechanisms as Bayesian games, and attempting to best-respond to perceived competition rather than playing the dominant strategy, then making beliefs more salient could amplify mistakes.

We test two levels of belief scaffolding, applied to both mechanisms.

\begin{itemize}
    \item \emph{First-Order Beliefs}: ``What do you expect the other players to do? Given their values are drawn randomly and they want to maximize their earnings, how do you think they will behave? Does this affect what you should do?''
    \item \emph{Second-Order Beliefs}: ``What do the other players think YOU will do? They know you are rational. They might try to anticipate your strategy. Does their belief about your behavior affect what they will do? And does that, in turn, affect what you should do?''
\end{itemize}

\begin{figure}[htbp]
    \centering
    \includegraphics[width=0.88\textwidth]{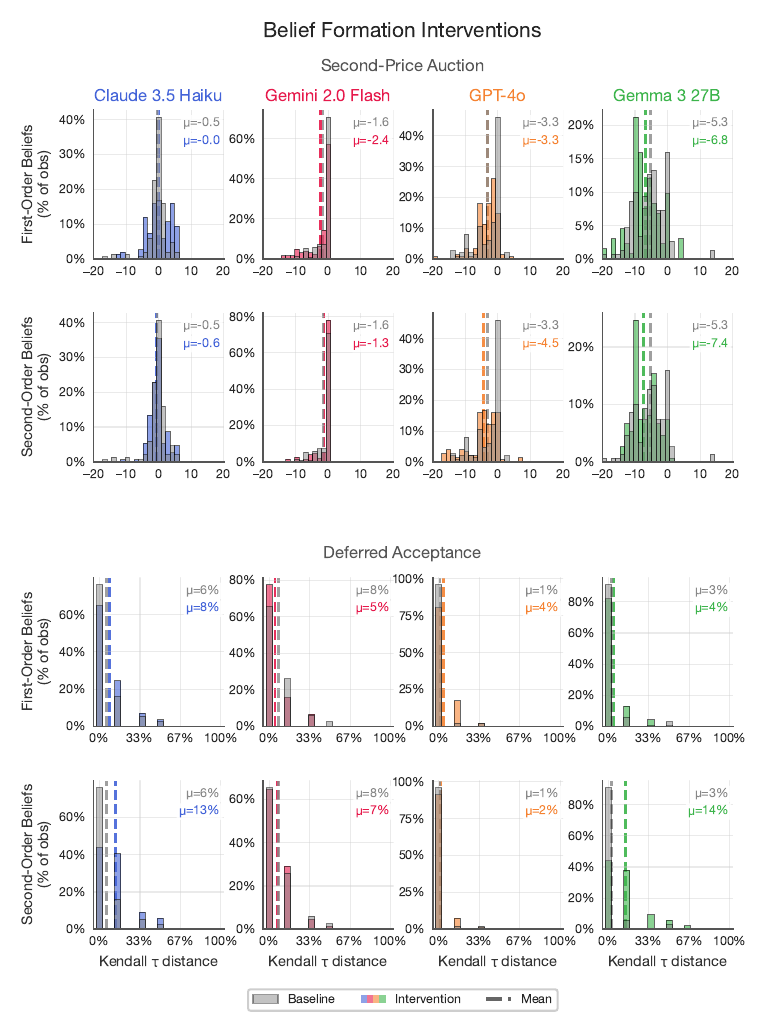}
    \caption{Belief formation interventions across both mechanisms. {\em Top two rows:} Second-price auction bid deviations under first-order and second-order belief scaffolds. {\em Bottom two rows:} Deferred acceptance Kendall $\tau$ errors under the same scaffolds. Baseline distributions in gray, interventions in color. Belief scaffolds almost uniformly worsen play, with second-order beliefs causing the most damage. Gemini is the notable exception, showing near-immunity to belief scaffolding.}
    \label{fig:beliefs}
\end{figure}

Indeed, we find that belief scaffolding almost uniformly worsens play across both mechanisms, and considering second-order beliefs hurts play marginally more. In auctions, \textit{Second-Order Beliefs} worsens the mean bid across models from $\mu = -2.67$ to $-3.47$; \textit{First-Order Beliefs} worsens it to $-3.04$. This effect is consistent for three of four models---only Gemini is unaffected. \rev{In absolute deviation the picture is sharper: under \textit{First-Order Beliefs} all four models, Gemini included, move farther from value, while under \textit{Second-Order Beliefs} Gemini alone moves closer (Section~\ref{sec:traces-probe}).} Models prompted to think about opponents shade further below value, undercutting perceived competition. \rev{The manipulation works as intended on the plans themselves: opponent-modeling language rises from $28\%$ of baseline plans to $43\%$ under first-order and $51\%$ under second-order prompts.} 

In DA, both interventions worsen play: \textit{First-Order Beliefs }moves from $\mu = 4.2\%$ to $5.1\%$, \textit{Second-Order Beliefs }to $9.1\%$. Again, three of four models get worse---Gemini is the exception, showing small improvements. The damage is particularly severe for Gemma, which moves from $2.6\%$ to $14.3\%$ under \textit{Second-Order Beliefs}.

Interestingly, Gemini is almost completely immune to belief scaffolding while the other tested models are hurt. In contrast, Gemma---which suffers the most severe damage under Second-Order Beliefs ($2.6\%$ to $14.3\%$ in DA)---appears most susceptible to this
failure mode. These results suggest that LLM reasoning styles are not monolithic; the heterogeneity across model families in susceptibility to belief-based reasoning may itself be relevant to mechanism design when markets serve diverse artificial agents.

\section{Additional Related Work} \label{sec:appendix-related}

\rev{The main text uses economic settings to study a broader question about supporting LLM decisions. This appendix explains the theories that motivate the interventions and relates the experiments to prior work on human and artificial agents.}

\paragraph{Complexity and Apparent Irrationality}
Recent work emphasizes the connection between strategic complexity and apparent irrationality. \citet{oprea2024decisions} argues that many anomalies attributed to non-standard preferences are better explained by computational complexity: subjects make mistakes not because their preferences are irrational but because the problem is hard. This framing aligns with our approach. We do not assume LLMs have ``biases'' in the behavioral economics sense (though we explore interventions probing this in the Appendix); rather, our primary question is whether the computational structure of strategic problems creates cognitive challenges and failure modes for LLMs.

\paragraph{Deferred Acceptance}

There is a deep literature in market design for both auctions and two-sided matching. 
For two-sided matching, \citet{gale1962college} introduced the DA algorithm, which produces stable matchings, and \citet{roth1982economics} showed that DA is strategy-proof for the proposing side.

The question of simplicity in DA was addressed first by \citet{ashlagi2018stable}, who show that DA is generally not OSP-implementable. However, under \textit{acyclic} priority structures---a condition introduced by \citet{ergin2002efficient}---the proposer-optimal stable matching rule can be implemented in an OSP manner (for the proposer). For unit-capacity schools, the relevant acyclicity condition rules out three students $a,b,c$ and two schools $s,t$ with $a \succ_s b \succ_s c$ and $c \succ_t a$. Under this condition, an OSP implementation of student-proposing DA exists.

\citet{gonczarowski2023strategyproofness} study descriptions of strategy-proof mechanisms that aim to make incentive properties transparent. They give a menu-based description for DA: the set of schools a student can obtain as it depends on others' reports, where the student's ranking is used to select within this menu. Such descriptions may help participants understand strategy-proofness without changing the extensive form---a complementary approach to OSP implementation. In an incentivized lab experiment, \citet{gonczarowski2024describing} find that a menu-based explanation significantly improved participants' measured understanding of DA's strategyproofness relative to standard descriptions, though average behavioral effects were modest.

\paragraph{Further Connections}
\rev{The following work places the interface comparisons in the broader literatures on simplicity and LLM decision making.}
\paragraph{Simple Mechanism Design}

 \citet{li2017obviously} formalized obvious strategy-proofness and showed that ascending clock auctions satisfy it while  SPSB auctions do not. Related notions include \textit{strategic simplicity} \citep{borgers2019strategically}, which requires that optimal play not depend on higher-order beliefs, and \textit{one-step simplicity} \citep{pycia2023theory}, which lets agents identify the prescribed action while considering at most one of their own future moves at a time. The SPSB auction satisfies strategic simplicity---truthful bidding is dominant regardless of beliefs about others---but is not OSP. Ascending clock auctions satisfy both. \citet{li2024designing} decomposes mechanism complexity into three cognitive dimensions: contingent reasoning, forward planning, and belief formation. We use this decomposition to structure our interventions.

\paragraph{LLMs as Economic Agents}

An emerging literature studies LLMs as simulated economic agents. \citet{horton2023large} introduces the ``homo silicus'' framework, showing that LLMs can replicate qualitative patterns from classic experiments in labor economics, bargaining, and social preferences. \citet{aher2023using} demonstrate that LLMs can reproduce human subject study results across a range of paradigms. \citet{zhu2024evidence} study LLMs in auction settings and find that OSP mechanisms improve play, a finding we replicate and extend in this paper.

In auction settings specifically, \citet{chen2023put} study LLMs in a multi-round auction environment and find substantial deviations from optimal play. \citet{fish2024algorithmic} examine whether LLMs can sustain collusive outcomes in repeated auctions. Our contribution differs in focus: we use the conceptual vocabulary of simple mechanism design---contingent reasoning, forward planning, belief formation---to probe and characterize the cognitive constraints that bind on LLM agents, and to ask \rev{whether the levers the same theory supplies for bounded human reasoners also move theirs---and whether behavioral improvement is accompanied by improvement in what the agents articulate}.

\begin{revblock}
\paragraph{Heuristics and the Faithfulness of Stated Reasoning}

Decision quality depends jointly on a procedure and the environment in which it is used \citep{simon1956rational}. Heuristics are selective rules that can work well in one setting and poorly in another \citep{gigerenzer2011heuristic, shah2008heuristics}. Our labels describe such rules in agents' stated plans: bidding below value to preserve a margin, for example, is appropriate to a first-price auction but misplaced in a second-price one. Because explanations need not faithfully report a model's computation \citep{turpin2023language, lanham2023measuring}, we evaluate these statements alongside realized bids rather than treating them as direct evidence of internal reasoning.
\end{revblock}